\documentclass[10pt,twocolumn,letterpaper]{article}

\usepackage{wacv}
\usepackage{times}
\usepackage{epsfig}
\usepackage{graphicx}
\usepackage{amsmath}
\usepackage{amssymb}
\usepackage{multirow}
\usepackage{array}
\usepackage{pifont}         

\newcommand{\cmark}{\ding{51}}
\newcommand{\xmark}{\ding{55}}

\usepackage[square,numbers]{natbib}

\def\wacvPaperID{806} 

\wacvfinalcopy 

\ifwacvfinal
\def\assignedStartPage{1} 
\fi

\ifwacvfinal
\usepackage[breaklinks=true,bookmarks=false]{hyperref}
\else
\usepackage[pagebackref=true,breaklinks=true,colorlinks,bookmarks=false]{hyperref}
\fi

\ifwacvfinal
\else
\fi

\begin{document}

\title{ImVoxelNet: Image to Voxels Projection for Monocular and Multi-View General-Purpose 3D Object Detection}

\author{
Danila Rukhovich\textsuperscript{1,2}, Anna Vorontsova\textsuperscript{1}, Anton Konushin\textsuperscript{1,2} \\
\textsuperscript{1}Samsung AI Center Moscow; \textsuperscript{2}Lomonosov Moscow State University \\
{\tt \small \{d.rukhovich,\ a.vorontsova,\ a.konushin\}@samsung.com}}

\maketitle

\begin{abstract}
In this paper, we introduce the task of multi-view RGB-based 3D object detection as an end-to-end optimization problem. To address this problem, we propose ImVoxelNet, a novel fully convolutional method of 3D object detection based on posed monocular or multi-view RGB images. The number of monocular images in each multi-view input can variate during training and inference; actually, this number might be unique for each multi-view input. ImVoxelNet successfully handles both indoor and outdoor scenes, which makes it general-purpose. Specifically, it achieves state-of-the-art results in car detection on KITTI (monocular) and nuScenes (multi-view) benchmarks among all methods that accept RGB images. Moreover, it surpasses existing RGB-based 3D object detection methods on the SUN RGB-D dataset. On ScanNet, ImVoxelNet sets a new benchmark for multi-view 3D object detection. The source code and the trained models are available at \url{https://github.com/saic-vul/imvoxelnet}.
\end{abstract}

\section{Introduction}

RGB images are an affordable and universal data source; therefore, RGB-based 3D object detection has been actively investigated in recent years. RGB images provide visual clues about the scene and its objects, yet they do not contain explicit information about the scene geometry and the absolute scale of the data. By virtue of that, detecting 3D objects from the RGB images is an ill-posed task. Given a monocular image, deep learning-based 3D object detection methods can only deduce the scale of the data. Moreover, the scene geometry cannot be unambiguously derived from the RGB images since some areas may be invisible. However, using several posed images might help obtain more information about the scene than a monocular RGB image. Accordingly, some 3D object detection methods \cite{simonelli2020disentangling, roddick2018orthographic} run multi-view inference. These methods obtain predictions on each monocular RGB image independently, then aggregate these predictions.

In contrast, we use multi-view inputs not only for inference but also for training. During both training and inference, the proposed method accepts posed multi-view inputs with an arbitrary number of views; this number might be unique for each multi-view input. Besides, our method can accept posed monocular inputs (treated as a special case of multi-view inputs). Furthermore, it works surprisingly well on monocular benchmarks.

All RGB-based 3D object detection methods are designed to be indoor or outdoor and work under certain assumptions about the scene and the objects. For instance, outdoor methods are typically evaluated on cars. In general, cars are of similar size, they are located on the ground, and their projections onto the Bird's Eye View (BEV) do not intersect. Accordingly, a BEV-plane projection contains much information on the 3D location of a car. So, a common approach in outdoor 3D object detection is to reduce a 3D object detection in a point cloud to a 2D object detection in the BEV plane. At the same time, indoor objects might have different heights and be randomly located in space, so their projections onto the floor plane provide little information about their 3D positions. Overall, the design of RGB-based 3D object detection methods tends to be domain-specific. 

To accumulate information from multiple inputs, we construct a voxel representation of the 3D space. We use this unified approach to detect objects in both indoor and outdoor scenes: we only choose between an indoor and outdoor head, while the meta-architecture remains the same.

In the proposed method, final predictions are obtained from 3D feature maps, which corresponds to the formulation of the point cloud-based detection problem. On this basis, we use off-the-shelf necks and heads from point cloud-based object detectors with no modifications. 

Our contribution is three-fold:
\begin{itemize}
    \item As far as we know, we are the first to formulate a task of end-to-end training for multi-view 3D object detection based on posed RGB images only.
    \item We propose a novel fully convolutional 3D object detector that works in both monocular and multi-view settings.
    \item With domain-specific heads, the proposed method achieves state-of-the-art results for both indoor and outdoor datasets.
\end{itemize}

\section{Related Works}

\subsection{Multi-view Scene Understanding}

Many scene understanding methods accept multi-view inputs. For instance, some scene understanding sub-tasks can only be solved given multi-view inputs. For example, the SLAM task implies reconstructing 3D scene geometry and estimating camera poses given a sequence of frames. Structure-from-Motion (SfM) approaches are designed to estimate camera poses and intrinsics from an unordered set of images, whereas Multi-View Stereo (MVS) methods use SfM outputs to build a 3D point cloud.

Other scene understanding sub-tasks might be reformulated to be multi-view. Several methods that use multi-view inputs to address these tasks have been proposed recently. For instance, 3D-SIS \cite{hou20193dsis} performs 3D instance segmentation based on a set of RGB-D inputs. MVPointNet \cite{jaritz2019mvpointnet} uses multi-view RGB-D inputs for 3D semantic segmentation. Atlas \cite{murez2020atlas} processes several monocular RGB images to perform 3D semantic segmentation and TSDF reconstruction jointly. 

\subsection{3D Object Detection.}

\textbf{Point cloud-based.} Point clouds are three-dimensional, so it seems natural to employ a 3D convolutional network for detection. However, this approach requires exhaustive computation that causes slow inference on large outdoor scenes. Recent outdoor methods \cite{yan2018second, lang2019pointpillars} decrease the runtime by projecting the 3D point cloud to the BEV plane. The common practice in point cloud processing is to subdivide a point cloud into voxels. The projection onto the BEV plane implies that all voxels in each vertical column should be encoded into a fixed-length feature map. Then, this pseudo-image can be passed to a 2D object detection network to obtain final predictions.

Indoor object detection methods generate object proposals for each point in a point cloud. However, some indoor objects are not convex, so the geometrical center of an indoor object may not belong to this object (e.g., the center of a table or a chair might be in between legs). Accordingly, an object proposal given by a single center point might be irrelevant, so indoor methods use deep Hough voting to generate proposals \cite{qi2019votenet, qi2020imvotenet, zhang2020h3dnet}. 

\textbf{Stereo-based.} Despite accepting more than one image, stereo-based methods cannot be considered multi-view as they use two images. In contrast, multi-view methods can process an arbitrary amount of inputs. Moreover, camera poses might be arbitrary for multi-view inputs, and for stereo inputs, the relative transformation between two cameras is known precisely and remains fixed while recording. This makes it possible to perform stereo reconstruction by estimating optical flow between the left and right images. Stereo-based methods rely heavily on the stereo assumptions, e. g., 3DOP \cite{chen20153dop} uses stereo reconstruction to generate object proposals, while TLNet \cite{qin2019triangulation} runs triangulation to merge proposals obtained for left and right images independently. Stereo R-CNN \cite{li2019stereorcnn} generates object proposals given both left and right images, then estimates object location by triangulating keypoints. 

\textbf{Monocular-based.} Mono3D \cite{chen2016mono3d} generates 3D anchors by aggregating clues from semantic maps, visible contours of the objects, and location priors via a complex energy function. Deep3DBox \cite{mousavian2017deep3dbox} uses discretization to estimate the orientation of each object and derives its 3D pose from constraints between 2D and 3D bounding boxes. MonoGRNet \cite{qin2019monogrnet} decomposes the 3D object detection problem into sub-tasks, namely object distance estimation, object location estimation, and object corners estimation. These sub-tasks are solved by separate networks, trained first stage-wise then altogether to refine 3D bounding boxes.

Other methods, e.g., \cite{chabot2017deepmanta, huang2018cooperative, nie2020total3dunderstanding}, exploit 2D detection and lift information from 2D to 3D. \cite{huang2018holistic, huang2018cooperative, nie2020total3dunderstanding} extend 2D detection network with a 3D branch that regresses object pose. Some methods make use of external data sources, e.g., DeepMANTA \cite{chabot2017deepmanta} uses an iterative coarse-to-fine algorithm of generating 2D object proposals, which are used to select a CAD model. 3D-RCNN \cite{kundu20183drcnn} also performs 2D detection and matches the outputs to 3D models. Then, it uses a render-and-compare approach to recover the shape and pose of an object. 

Monocular indoor 3D object detection is a less explored problem, with only SUN RGB-D \cite{song2015sunrgbd} benchmark existing. This benchmark implies that indoor 3D object detection is a sub-task of total scene understanding. Beside detecting 3D objects, \cite{huang2018holistic, huang2018cooperative, nie2020total3dunderstanding} estimate camera poses and room layouts. The most recent Total3DUnderstanding \cite{nie2020total3dunderstanding} reconstructs object meshes using an attention mechanism to consider relationships between objects.

Some outdoor 3D object detection methods \cite{simonelli2020disentangling, roddick2018orthographic} are evaluated on the nuScenes \cite{caesar2020nuscenes} dataset on multi-view inputs. Specifically, these methods infer on each monocular RGB image, then aggregate the outputs. Aggregation is an inevitable part of the pipeline; however, doing this on the latest stage is controversial, as spatial information might not be exploited as effectively as possible. 

So, none of the existing methods formulate 3D object detection given multiple RGB images as an end-to-end optimization problem.

\begin{figure*}
    \centering
        \includegraphics[width=0.95\linewidth]{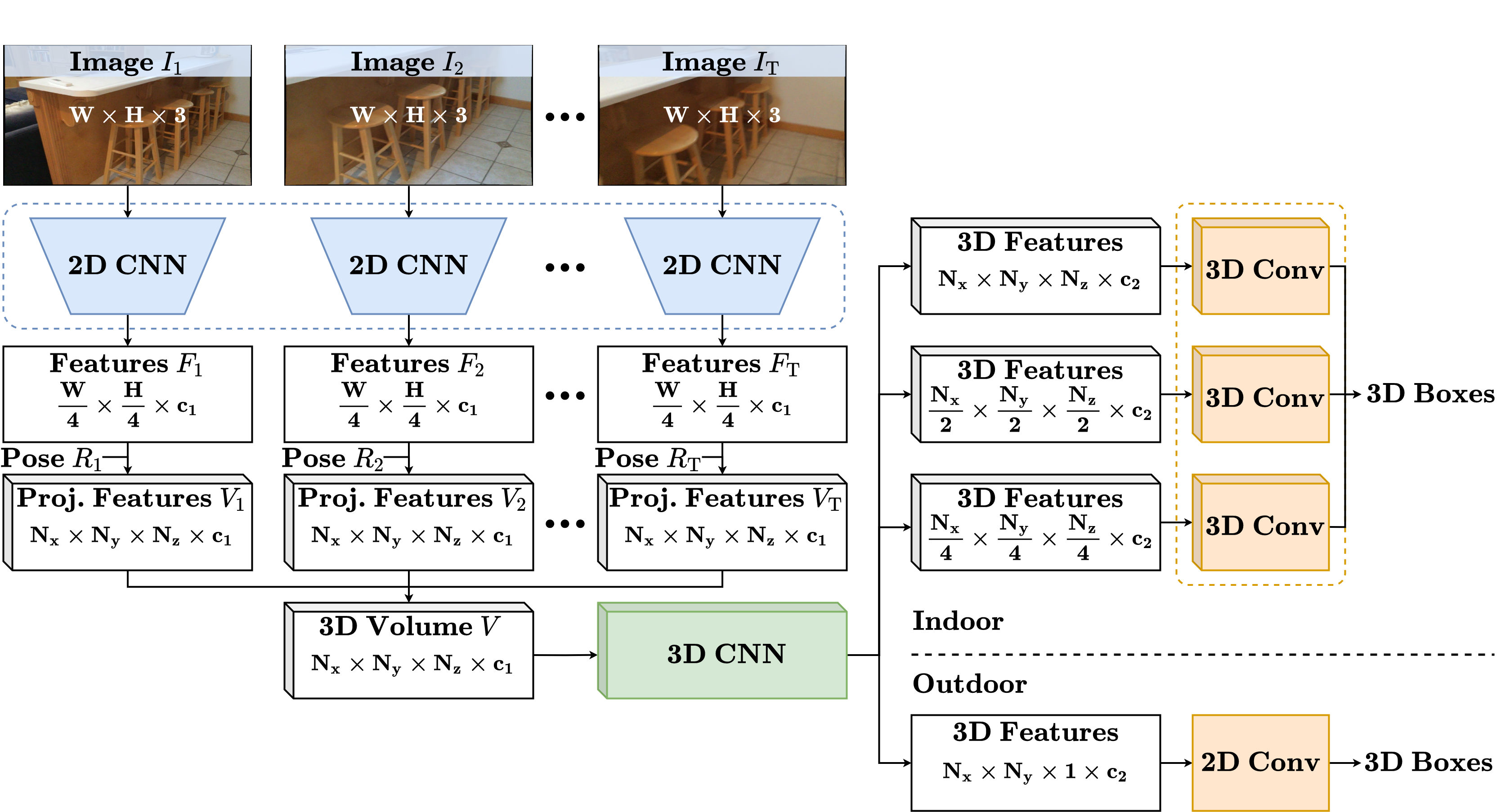}
    \caption{The general scheme of the proposed ImVoxelNet. Dashed lines around network blocks denote that network weights are shared across multiple inputs.}
    \label{fig:scheme}
\end{figure*}

\section{Proposed Method}

Our method accepts an arbitrary-sized set of RGB inputs along with camera poses. First, we extract features from the given images using a 2D convolutional backbone. Then, we project the obtained image features to a 3D voxel volume. For each voxel, the projected features from several images are aggregated via a simple element-wise averaging. Next, the voxel volume with assigned features is passed to a 3D convolutional network referred to as \textit{neck}. The outputs of the neck serve as inputs to the last few convolutional layers (\textit{head}) that predict bounding box features for each anchor. The resulting bounding boxes are parameterized as $(x, y, z, w, h, l, \theta)$, where $(x, y, z)$ are the coordinates of the center, $w, h, l$ are for width, height, and length, and $\theta$ is the rotation angle around $z$-axis. The general scheme of the proposed method is depicted in Fig. \ref{fig:scheme}.

2D features projection and 3D neck network have been proposed in \cite{murez2020atlas, hou20193dsis}. First, we briefly outline these steps. Then, we introduce a novel multi-scale 3D head designed for indoor detection.

\subsection{3D Volume Construction}

Let $I_t \in \mathbb{R}^{W \times H \times 3}$ be the $t$-th image in a set of $T$ images. Here, $T > 1$ in case of multi-view inputs and $T = 1$ for single-view inputs. Following \cite{murez2020atlas}, we first extract 2D features from passed inputs using a pretrained 2D backbone. It outputs four feature maps of shapes $\frac{W}{4} \times \frac{H}{4} \times c_0$, $\frac{W}{8} \times \frac{H}{8} \times 2c_0$, $\frac{W}{16} \times \frac{H}{16} \times 4c_0$, and $\frac{W}{32} \times \frac{H}{32} \times 8c_0$. We aggregate the obtained feature maps via Feature Pyramid Network (FPN), which outputs one tensor $F_t$ of shape $\frac{W}{4} \times \frac{H}{4} \times c_1$. $c_0$ and $c_1$ are backbone-specific; actual values are present in \ref{sec:implementation}.

For t-th input, the extracted 2D features $F_t$ are then projected into a 3D voxel volume $V_t \in \mathbb{R}^{N_x \times N_y \times N_z \times c_1}$. We set the $z$-axis to be perpendicular to the floor plane, with the $x$-axis pointing forward and the $y$-axis being orthogonal to both $x$ and $z$-axes. For each dataset, there are known spatial limits for all three axes, estimated empirically in \cite{zhang2020h3dnet, lang2019pointpillars, murez2020atlas}. Let us denote these limits as $x_\mathsf{min}, x_\mathsf{max}, y_\mathsf{min}, y_\mathsf{max}, z_\mathsf{min}, z_\mathsf{max}$. For a fixed voxel size $s$, spatial constraints can be formulated as $N_x s = x_\mathsf{max} - x_\mathsf{min}$, $N_y s = y_\mathsf{max} - y_\mathsf{min}$, and $N_z s = z_\mathsf{max} - z_\mathsf{min}$. We use a pinhole camera model, which determines the correspondence between 2D coordinates $(u, v)$ in feature map $F_t$ and 3D coordinates $(x, y, z)$ in volume $V_t$:
\[
\left[\begin{array}{c} u \\ v \end{array}\right] =
\Pi
\left[ \begin{array}{ccc}
\frac{1}{4} & 0 & 0 \\
0 & \frac{1}{4} & 0 \\
0 & 0 & 1
\end{array} \right] K R_t
\left[\begin{array}{c} x \\ y \\ z \\ 1 \end{array}\right],
\]
where $K$ and $R_t$ are the intrinsic and extrinsic matrices, and $\Pi$ is a perspective mapping. After projecting 2D features, all voxels along a camera ray get filled with the same features. We also define a binary mask $M_t$ of the same shape as $V_t$, which indicates whether each voxel is inside the camera frustum. Thus, for each image $I_t$, the mask $M_t$ is defined as:
\[M_t(x, y, z) = \begin{cases}
    1, & \text{if}\ 0 \le u < \frac{W}{4}\ \text{and}\ 0 \le v < \frac{H}{4} \\
    0, & \text{otherwise}.
\end{cases}\]
Then, we project $F_t$ for each valid voxel in a volume $V_t$:
\[V_t(x, y, z) = \begin{cases}
    F_t(u, v), & \text{if}\ M_t(x, y, z) = 1 \\
    0, & \text{otherwise}.
\end{cases}\]
The aggregated binary mask $M$ is a sum of $M_1, \dots, M_t$:
\[M(x, y, z) = \begin{cases}
    \sum_t M_t(x, y, z), & \text{if}\ \sum_t M_t(x, y, z) > 0 \\
    1, & \text{otherwise}.
\end{cases}\]
Finally, we obtain the 3D volume $V$ by averaging projected features in volumes $V_1, \dots, V_t$ across valid voxels:
\[V = \frac{1}{M} \sum_t M_t V_t.\]

\subsection{3D Feature Extraction} \label{sec:features}

\textbf{Indoor.} Following \cite{murez2020atlas, hou20193dsis}, we pass the voxel volume $V$ through a 3D convolutional encoder-decoder network to refine the features. For indoor scenes, we use an encoder-decoder architecture from \cite{murez2020atlas}. However, with over 48 3D convolutional layers, the original network is computationally heavy and slow on inference. For a better performance, we simplify the network by reducing the number of time-consuming 3D convolutional layers. The simplified encoder has only three downsampling residual blocks, each with three 3D convolutional layers. The simplified decoder consists of three upsampling blocks, and each upsampling block is made up with a transposed 3D convolutional layer with stride 2 followed by another 3D convolutional layer. The decoder branch outputs three feature maps of the following shapes: $\frac{N_x}{4} \times \frac{N_y}{4} \times \frac{N_z}{4} \times c_2$, $\frac{N_x}{2} \times \frac{N_y}{2} \times \frac{N_z}{2} \times c_2$, and $N_x \times N_y \times N_z \times c_2$. For the actual value of $c_2$, see \ref{sec:implementation}.

\textbf{Outdoor.} Outdoor methods \cite{sindagi2019mvxnet, lang2019pointpillars, yan2018second} reduce 3D object detection in 3D space to 2D object detection in the BEV plane. In these methods, both the neck and head are composed of 2D convolutions. The outdoor head accepts a 2D feature map, so we should obtain a 2D representation of a constructed 3D voxel volume to use in our method. In order to do that, we use the encoder part of the encoder-decoder architecture from \cite{murez2020atlas}. After passing through several 3D convolutional and downsampling layers of this encoder, a voxel volume $V$ of shape $N_x \times N_y \times N_z \times c_1$ is mapped to the tensor of shape $N_x \times N_y \times c_2$.

\subsection{Detection Heads}

ImVoxelNet constructs a 3D voxel representation of the space; thus, it can use the head from point cloud-based 3D object detection methods. Therefore, instead of time-consuming custom architecture implementation, one can employ state-of-the-art methods with no modifications. However, the design of heads significantly differs for outdoor \cite{lang2019pointpillars, yan2018second} and indoor \cite{qi2019votenet, qi2020imvotenet} methods.

\subsubsection{Outdoor Head} \label{sec:outdoor}

We reformulate outdoor 3D object detection as 2D object detection in the BEV plane following the common practice. We use the 2D anchor head that appeared to be efficient \cite{lang2019pointpillars, yan2018second} on KITTI \cite{geiger2012kitti} and nuScenes \cite{caesar2020nuscenes} datasets. Since outdoor 3D detection methods are evaluated on cars, all objects are of a similar scale and belong to the same category. For single-scale and single-class detection, the head consists of two parallel 2D convolutional layers. One layer estimates class probability, while the other regresses seven parameters of the bounding box.

\textbf{Input.} The input is a tensor of shape $N_x \times N_y \times c_2$.

\textbf{Output.} For each 2D BEV anchor, the head returns a class probability $p$ and a 3D bounding box as a 7-tuple:
\[\Delta x = \frac{x^\mathsf{gt}-x^\mathsf{a}}{d^\mathsf{a}}, \Delta y = \frac{y^\mathsf{gt}-y^\mathsf{a}}{d^\mathsf{a}}, \Delta z = \frac{z^\mathsf{gt}-z^\mathsf{a}}{d^\mathsf{a}},\]
\[\Delta w = \log \frac{w^\mathsf{gt}}{w^\mathsf{a}}, \Delta l = \log \frac{l^\mathsf{gt}}{l^\mathsf{a}}, \Delta h = \log \frac{h^\mathsf{gt}}{h^\mathsf{a}},\]
\[ \Delta \theta=\sin(\theta^\mathsf{gt}-\theta^\mathsf{a}).\]
Here $\cdot^\mathsf{gt}$ and $\cdot^\mathsf{a}$ are the ground truth and anchor boxes, respectively. The length of the bounding box diagonal $d^\mathsf{a}=\sqrt{{(w^\mathsf{a})}^2+{(l^\mathsf{a})}^2}$. $z_\mathsf{a}$ is constant for all anchors since they are located in the BEV plane.

\textbf{Loss.} We use the loss function introduced in SECOND \cite{yan2018second}. The total outdoor loss consists of several loss terms, namely smooth mean absolute error as a location loss $L_\mathsf{loc}$, focal loss for classification $L_\mathsf{cls}$, and cross-entropy loss for direction $L_\mathsf{dir}$. Overall, we can formulate the outdoor loss as \[L_\mathsf{outdoor}=\frac{1}{n_\mathsf{pos}}(\lambda_\mathsf{loc}L_\mathsf{loc} + \lambda_\mathsf{cls}L_\mathsf{cls} + \lambda_\mathsf{dir}L_\mathsf{dir}),\]
where $n_\mathsf{pos}$ is the number of positive anchors, $\lambda_\mathsf{loc}=2$, $\lambda_\mathsf{cls}=1$, $\lambda_\mathsf{dir}=0.2$.

\subsubsection{Indoor Head} \label{sec:indoor}

All modern indoor 3D object detection methods \cite{qi2019votenet, qi2020imvotenet, zhang2020h3dnet} perform deep Hough voting for sparse point cloud representation. In contrast, we follow \cite{murez2020atlas, hou20193dsis} and use dense voxel representation of intermediate features. To the best of our knowledge, there is no dense 3D multi-scale head for 3D object detection. We construct such a head inspired by a 2D detection method FCOS \cite{tian2019fcos}. An original FCOS head accepts 2D features from FPN and estimates 2D bounding boxes via 2D convolutional layers. To adapt FCOS for 3D detection, we replace 2D convolutions with 3D convolutions to process 3D inputs. Following FCOS and ATSS \cite{zhang2020atss}, we apply center sampling to select candidate object locations. In these works, 9 ($3 \times 3$) candidates were chosen; since we operate in 3D space, we set a limit of 27 candidate locations per object ($3 \times 3 \times 3$). The resulting head consists of three 3D convolutional layers for classification, location, and centerness, respectively, with weights shared across all object scales.

\textbf{Input.} A multi-scale input is composed of three tensors of shapes $\frac{N_x}{4} \times \frac{N_y}{4} \times \frac{N_z}{4} \times c_2$, $\frac{N_x}{2} \times \frac{N_y}{2} \times \frac{N_z}{2} \times c_2$, and $N_x \times N_y \times N_z \times c_2$. 

\textbf{Output.} For each 3D location $(x^\mathsf{a}, y^\mathsf{a}, z^\mathsf{a})$ and each of three scales, the head estimates a class probability $p$, a centerness $c$, and a 3D bounding box as a 7-tuple:
\[\Delta x_\mathsf{min}=x_\mathsf{min}^\mathsf{gt}-x^\mathsf{a}, \Delta x_\mathsf{max}=x_\mathsf{max}^\mathsf{gt}-x^\mathsf{a},\]
\[ \Delta y_\mathsf{min}=y_\mathsf{min}^\mathsf{gt}-y^\mathsf{a}, \Delta y_\mathsf{max}=y_\mathsf{max}^\mathsf{gt}-y^\mathsf{a},\]
\[ \Delta z_\mathsf{min}=z_\mathsf{min}^\mathsf{gt}-z^\mathsf{a}, \Delta z_\mathsf{max}=z_\mathsf{max}^\mathsf{gt}-z^\mathsf{a},\theta.\]
Here, $x_\mathsf{min}^\mathsf{gt}, x_\mathsf{max}^\mathsf{gt}, y_\mathsf{min}^\mathsf{gt}, y_\mathsf{max}^\mathsf{gt}, z_\mathsf{min}^\mathsf{gt}, z_\mathsf{max}^\mathsf{gt}$ denote the minimum and maximum coordinates along axes of a ground truth bounding box.

\textbf{Loss.} We adapt the loss function used in the original FCOS \cite{tian2019fcos}. It consists of focal loss for classification $L_\mathsf{cls}$, cross-entropy loss for centerness $L_\mathsf{cntr}$, and IoU loss for location $L_\mathsf{loc}$. Since we address the 3D detection task instead of the 2D detection task, we replace 2D IoU loss with rotated 3D IoU loss \cite{zhou2019iou}. In addition, we update ground truth centerness with the third dimension.
The resulting indoor loss can be written as \[L_\mathsf{indoor}=\frac{1}{n_\mathsf{pos}}(L_\mathsf{loc} + L_\mathsf{cls} + L_\mathsf{cntr}),\]
where $n_\mathsf{pos}$ is the number of positive 3D locations.

\subsection{Extra 2D Head}

In some indoor benchmarks, the 3D object detection task is formulated as a sub-task of scene understanding. Accordingly, evaluation protocols imply solving various scene understanding tasks rather than only estimating 3D bounding boxes. Following \cite{huang2018holistic, huang2018cooperative, nie2020total3dunderstanding}, we predict camera rotations and room layouts. Similar to \cite{nie2020total3dunderstanding}, we add a simple head for joint $R_t$ and 3D layout estimation. This extra head consists of two parallel branches: two fully connected layers output room layout and the other two fully connected layers estimate camera rotation.

\textbf{Input.} The input is a single tensor of shape $8c_0$, obtained through global average pooling of the backbone output.

\textbf{Output.} The head outputs camera pose as a tuple of pitch $\beta$ and roll $\gamma$ and a 3D layout box as a 7-tuple $(x, y, z, w, l, h, \theta)$. As \cite{nie2020total3dunderstanding}, we set yaw angle and shift to zeros.

\textbf{Loss.} We modify losses used in \cite{nie2020total3dunderstanding} to make them consistent with the losses used to train a detection head. Accordingly, we define layout loss $L_\mathsf{layout}$ as rotated 3D IoU loss between predicted and ground truth layout boxes; this is the same loss as we use in \ref{sec:indoor}. For camera rotation estimation, we use $L_\mathsf{pose} = |\sin(\beta^\mathsf{gt}-\beta)|+|\sin(\gamma^\mathsf{gt}-\gamma)|$ similar to \ref{sec:outdoor}. Overall, the extra loss can be formulated as
\[L_\mathsf{extra}=\lambda_\mathsf{layout}L_\mathsf{layout}+\lambda_\mathsf{pose}L_\mathsf{pose},\]
where $\lambda_\mathsf{layout}=0.1$ and $\lambda_\mathsf{pose}=1.0$.

\section{Experiments}

\subsection{Datasets}

We evaluate the proposed method on four datasets: indoor ScanNet \cite{dai2017scannet} and SUN RGB-D \cite{song2015sunrgbd}, and outdoor KITTI \cite{geiger2012kitti} and nuScenes \cite{caesar2020nuscenes}. SUN RGB-D and KITTI are benchmarked in monocular mode, while for ScanNet and nuScenes, we address the detection problem in multi-view formulation.

\textbf{KITTI.} The KITTI object detection dataset \cite{geiger2012kitti} is the most decisive outdoor benchmark for monocular 3D object detection. It consists of 3711 training, 3768 validation and 7518 test images. The common practice \cite{simonelli2020disentangling, liu2020smoke} is to report results on validation subset and submit test predictions to an open leaderboard. All 3D object annotations have a difficulty level: \textit{easy}, \textit{moderate}, and \textit{hard}. A 3D object detection method is assessed according to the results on \textit{moderate} objects from the test set. Following \cite{simonelli2020disentangling, liu2020smoke}, we evaluate our method only on objects of the \textit{car} category.

\textbf{nuScenes.} The nuScenes dataset \cite{caesar2020nuscenes} provides data for developing algorithms addressing self-driving-related tasks. It contains LiDAR point clouds, RGB images captured by six cameras, accompanied by IMU and GPS measurements. The dataset covers 1000 video sequences, each recorded for 20 seconds, totalling 1.4 million images and 390 000 point clouds. Training split covers 28 130 scenes, and validation split contains 6019 scenes. The annotation contains 1.4 million objects divided into 23 categories. Following \cite{simonelli2020disentangling}, the accuracy of 3D detection is measured only on \textit{car} category. In this benchmark, not only the average precision (AP) metric but average translation error (ATE), average scale error (ASE), and average orientation error (AOE) are calculated as well.

\begin{table}[!ht]
    \centering \small
    \begingroup \setlength{\tabcolsep}{2pt}
    \begin{tabular}{l|cccccc|c}
    \hline
    Dataset & $x_\mathsf{min}$ & $x_\mathsf{max}$ & $y_\mathsf{min}$ & $y_\mathsf{max}$ & $z_\mathsf{min}$ & $z_\mathsf{max}$ & $s$ \\ \hline
    KITTI & -39.68 & 39.68 & 0 & 69.12 & -2.92 & 0.92 & \multirow[c]{2}{*}{0.32} \\
    nuScenes & -49.92 & 49.92 & -49.92 & 49.92 & -2.92 & 0.92 & \\
    SUN RGB-D & -3.2 & 3.2 & 0 & 6.4 & -2.28 & 0.28 & \multirow[c]{2}{*}{0.16} \\
    ScanNet & -3.2 & 3.2 & -3.2 & 3.2 & -1.28 & 1.28 & \\ \hline
    \end{tabular} \endgroup
    \caption{Implementation details. Axis limits and voxel size $s$ are measured in meters.}
    \label{tab:implementation}
\end{table}

\begin{table*}[!h]
    \centering \small
    \begingroup \setlength{\tabcolsep}{2pt}
    \begin{tabular}{l|c|ccc|ccc}
    \hline
    \multirow[c]{2}{*}{Method} & \multirow[c]{2}{*}{Depth} & \multicolumn{3}{c|}{AP\textsubscript{3D}@0.7 (\emph{val}/\emph{test})} & \multicolumn{3}{c}{AP\textsubscript{BEV}@0.7 (\emph{val}/\emph{test})} \\
    & & Easy & Moderate & Hard & Easy & Moderate & Hard \\ \hline
    MonoFENet\cite{bao2019monofenet} & \cmark & 17.54 / \phantom{0}8.34 & 11.16 / \phantom{0}5.14 & \phantom{0}9.74 / 4.10 & 30.21 / 17.03 & 20.47 / 11.03 & 17.58 / \phantom{0}9.05 \\
    AM3D\cite{ma2019am3d} & \cmark & 32.23 / 16.50 & 21.09 / 10.74 & 17.26 / 9.52 & 43.75 / 25.03 & 28.39 / 17.32 & 23.87 / 14.91 \\
    D4LCN\cite{ding2020d4lcn} & \cmark & 26.97 / 16.65 & 21.71 / 11.72 & 18.22 / 9.51 & 34.82 / 22.51 & 25.83 / 16.02 & 23.53 / 12.55 \\ \hline
    OFTNet\cite{roddick2018orthographic} & \xmark & \phantom{0}4.47 / \phantom{0}1.32 & \phantom{0}3.27 / \phantom{0}1.61 & \phantom{0}3.29 / \phantom{0}1.00 & 11.06 / \phantom{0}7.16 & \phantom{0}8.79 / \phantom{0}5.69 &  \phantom{0}8.91 / \phantom{0}4.61 \\
    GS3D\cite{li2019gs3d} & \xmark & 13.46 / \phantom{0}4.47 & 10.97 / \phantom{0}2.90 & 10.38 / 2.47 &  \phantom{00.0}-- / \phantom{0}8.41 &  \phantom{00.0}-- / \phantom{0}6.08 &  \phantom{00.0}-- / \phantom{0}4.94 \\
    MonoGRNet\cite{qin2019monogrnet} & \xmark & 13.88 / \phantom{0}9.61 & 10.19 / \phantom{0}5.74 & \phantom{0}7.62 / 4.25 & \phantom{00.0}-- / 18.19 & \phantom{00.0}-- / 11.17 & \phantom{00.0}-- / \phantom{0}8.73 \\
    MonoDIS\cite{simonelli2020disentangling} & \xmark & 18.05 / 10.37 & 14.98 / \phantom{0}7.94 & 13.42 / 6.40 & 24.26 / 17.23 & 18.43 / 13.19 & 16.95 / 11.12 \\
    SMOKE\cite{liu2020smoke} & \xmark & 14.76 / 14.03 & 12.85 / \phantom{0}9.76 & 11.50 / 7.84 & 19.99 / 20.83 & 15.61 / 14.49 & 15.28 / 12.75 \\
    M3D-RPN\cite{brazil2019m3drpn} & \xmark & 20.27 / 14.76 & 17.06 / \phantom{0}9.71 & 15.21 / 7.42 & 25.94 / 21.02 & 21.18 / 13.67 & 17.90 / 10.23 \\
    RTM3D\cite{li2020rtm3d} & \xmark & 20.77 / 14.41 & 16.86 / 10.34 & \textbf{16.63} / 8.77 & 25.56 / 19.17 & 22.12 / 14.20 & \textbf{20.91} / 11.99 \\
    ImVoxelNet & \xmark & \textbf{24.54} / \textbf{17.15} & \textbf{17.80} / \textbf{10.97} & 15.67 / \textbf{9.15} & \textbf{31.67} / \textbf{25.19} & \textbf{23.68} / \textbf{16.37} & 19.73 / \textbf{13.58} \\ \hline
    \end{tabular} \endgroup
    \caption{Scores for \textit{car} category on the KITTI dataset. The \textit{depth} column indicates whether this modality is used for training.}
    \label{tab:kitti}
\end{table*}

\begin{table*}[ht!]
    \centering \small
    \begingroup \setlength{\tabcolsep}{2pt}
    \begin{tabular}{l|cc|ccccc|ccc}
        \hline
        \multirow[c]{2}{*}{Method} & \multirow[c]{2}{*}{RGB}  & \multirow[c]{2}{*}{PC} & \multicolumn{5}{c|}{AP$\uparrow$[\%]} & \multicolumn{3}{c}{TP$\downarrow$} \\
        & & & 0.5m & 1.0m & 2.0m & 4.0m & mean & ATE [m] & ASE[1-IoU] & AOE[rad] \\ \hline
        PointPillar\cite{lang2019pointpillars} & \xmark & \cmark & 55.5 & 71.8 & 76.1 & 78.6 & 70.5 & 0.27 & 0.17 & 0.19 \\ \hline
        OFTNet \cite{roddick2018orthographic, simonelli2020disentangling} & \cmark & \xmark & -- & -- & 27.0 & -- & -- & 0.65 & 0.16 & 0.18 \\
        MonoDIS \cite{simonelli2020disentangling} & \cmark & \xmark & 10.7 & 37.5 & \textbf{69.0} & \textbf{85.7} & 50.7 & 0.61 & \textbf{0.15} & \textbf{0.08} \\
        ImVoxelNet & \cmark & \xmark & \textbf{19.3} & \textbf{44.8} & 66.3 & 77.0 & \textbf{51.8} & \textbf{0.52} & \textbf{0.15} & \textbf{0.08} \\ \hline
    \end{tabular} \endgroup
    \caption{Scores for \textit{car} category on the nuScenes dataset. The RGB and PC columns indicate data modalities used for both training and inference.}
    \label{tab:nuscenes}
\end{table*}

\textbf{SUN RGB-D.} SUN RGB-D \cite{song2015sunrgbd} is one of the first and most well-known indoor 3D datasets. It contains 10 335 images captured in various indoor places alongside corresponding depth maps obtained with four different sensors and camera poses. The training split is composed of 5285 frames, while the rest 5050 frames comprise the validation subset. The annotation includes 58 657 objects. For each frame, a room layout is provided.

\textbf{ScanNet.} The ScanNet dataset \cite{dai2017scannet} contains 1513 scans covering over 700 unique indoor scenes, out of which 1201 scans belong to a training split, and 312 scans are used for validation. Overall, this dataset contains over 2.5 million images with corresponding depth maps and camera poses, alongside reconstructed point clouds with 3D semantic annotation. We estimate 3D bounding boxes from semantic point clouds following the standard protocol \cite{qi2019votenet}. The resulting object bounding boxes are axis-aligned, so we do not predict the rotation angle $\theta$ for ScanNet.

\subsection{Implementation Details} \label{sec:implementation}

\textbf{3D Volume.} We use ResNet-50 \cite{he2016resnet} as a feature extractor. Accordingly, the number of convolutions in the first convolutional block $c_0$ equals 256. We set both the 3D volume feature size $c_1$ and the ouput feature size $c_2$ to 256 as proposed in \cite{lang2019pointpillars, yan2018second}.

Indoor and outdoor scenes are of different absolute scales. Therefore, we choose the spatial sizes of the feature volume for each dataset considering the data domain. We use the values provided in previous works \cite{murez2020atlas, lang2019pointpillars, yan2018second, sindagi2019mvxnet}, as shown in Tab. \ref{tab:implementation}. Thus, using anchor settings of the 3D head in \cite{lang2019pointpillars, sindagi2019mvxnet}, we set voxel size $s$ as $0.32$ meters for outdoor datasets. Minimal and maximal values for all three axes for outdoor datasets also follow the point cloud ranges for \textit{car} class in \cite{lang2019pointpillars, sindagi2019mvxnet}. For selecting indoor dataset constraints we follow \cite{murez2020atlas}, where the room size is $6.4 \times 6.4 \times 2.56$ meters. The only change is that we are increasing voxels size $s$ from $0.04$ to $0.16$ to increase memory efficiency. 

\textbf{Training.} During training, we optimize $L_\mathsf{indoor}$ for indoor datasets and $L_\mathsf{outdoor}$ for outdoor datasets, unless told otherwise. We use Adam optimizer with an initial learning rate set to $0.0001$ and weight decay of $0.0001$. The implementation is based on the MMDetection framework \cite{chen2019mmdetection} and uses its default training settings. The network is trained for 12 epochs, and the learning rate is reduced by ten times after the 8th and 11th epoch. For ScanNet, SUN RGB-D, and KITTI, the network sees each scene three times every training epoch. We use 8 Nvidia Tesla P40 GPUs for training, distributing one scene (multi-view scenario) or four images (monocular scenario) per GPU. We randomly apply horizontal flip and resize inputs in monocular experiments by no more than 25\% of their original resolution. Moreover, in indoor scenes, we can augment 3D voxel representations similar to point cloud-based methods, so we randomly shift a voxel grid center by at most 1m along each axis.

\textbf{Inference.} During inference, outputs are filtered with a Rotated NMS algorithm, which is applied to objects projections onto the ground plane.

\begin{table*}
    \centering \small
    \begingroup \setlength{\tabcolsep}{2pt}
    \begin{tabular}{l|cccccccccc|c||ccc}
    \hline
    Method & bed & chair & sofa & table & desk & dresser & nstand & sink & cabinet & lamp & mAP &  Layout$\uparrow$[IoU] & Pitch$\downarrow$[\textdegree] & Roll$\downarrow$[\textdegree]\\ \hline
    3DGP\cite{choi20133dgp} & \phantom{0}5.62 & \phantom{0}2.31 & \phantom{0}3.24 & \phantom{0}1.23 & -- & -- & -- & -- & -- & -- & -- & 19.2 & -- & --\\
    HoPR\cite{huang2018holistic} & 58.29 & 13.56 & 28.37 & 12.12 & \phantom{0}4.79 & 13.71 & \phantom{0}8.80 & \phantom{0}2.18 & \phantom{0}0.48 & \phantom{0}2.41 & 14.47 & 54.9 & 7.60 & 3.12 \\
    CooP\cite{huang2018cooperative} & 63.58 & 17.12 & 41.22 & 26.21 & \phantom{0}9.55 & \phantom{0}4.28 & \phantom{0}6.34 & \phantom{0}5.34 & \phantom{0}2.63 & \phantom{0}1.75 & 17.80 & 56.9 & 3.28 & 2.19 \\
    T3DU\cite{nie2020total3dunderstanding} & 59.03 & 15.98 & 43.95 & 35.28 & 23.65 & 19.20 & \phantom{0}6.87 & 14.40 & 11.39 & \phantom{0}3.46 & 23.32 & 57.6 & 3.68 & 2.59 \\
    ImVoxelNet & \textbf{79.17} & \textbf{63.07} & \textbf{60.59} & \textbf{51.14} & \textbf{31.20} & \textbf{35.45} & \textbf{38.38} & \textbf{45.12} & \textbf{19.24} & \textbf{13.27} & \textbf{43.66} & \textbf{59.3} & \textbf{2.63} & \textbf{1.96} \\ \hline
    \end{tabular} \endgroup
    \caption{AP@0.15 scores for 10 out of 37 object categories \cite{nie2020total3dunderstanding} from the SUN RGB-D dataset, alongside room layout and camera pose estimation metrics.}
    \label{tab:sunrgbd_total_10}
\end{table*}

\begin{table*}
    \centering \small
    \begingroup \setlength{\tabcolsep}{2pt}
    \begin{tabular}{l|cc|cccccccccccccccccc|c}
    \hline
    Method & RGB & PC & cab & bed & chair & sofa & tabl & door & wind & bkshf & pic & cntr & desk & curt & fridg & showr & toil & sink & bath & ofurn & mAP \\ \hline
    3D-SIS\cite{hou20193dsis} & \xmark & \cmark & 12.8 & 63.1 & 66.0 & 46.3 & 26.9 & \phantom{0}8.0 & \phantom{0}2.8 & \phantom{0}2.3 & \phantom{0}0.0 & \phantom{0}6.9 & 33.3 & \phantom{0}2.5 & 10.4 & 12.2 & 74.5 & 22.9 & 58.7 & \phantom{0}7.1 & 25.4 \\
    3D-SIS\cite{hou20193dsis} & \cmark & \cmark & 19.8 & 69.7 & 66.2 & 71.8 & 36.1 & 30.6 & 10.9 & 27.3 & \phantom{0}0.0 & 10.0 & 46.9 & 14.1 & 53.8 & 36.0 & 87.6 & 43.0 & 84.3 & 16.2 & 40.2 \\
    VoteNet\cite{qi2019votenet} & \xmark & \cmark & 36.3 & 87.9 & 88.7 & 89.6 & 58.8 & 47.3 & 38.1 & 44.6 & \phantom{0}7.8 & 56.1 & 71.7 & 47.2 & 45.4 & 57.1 & 94.9 & 54.7 & 92.1 & 37.2 & 58.7 \\
    H3DNet\cite{zhang2020h3dnet} & \xmark & \cmark & \textbf{49.4} & \textbf{88.6} & \textbf{91.8} & \textbf{90.2} & \textbf{64.9} & \textbf{61.0} & \textbf{51.9} & \textbf{54.9} & \textbf{18.6} & \textbf{62.0} & \textbf{75.9} & \textbf{57.3} & 57.2 & \textbf{75.3} & \textbf{97.9} & \textbf{67.4} & \textbf{92.5} & \textbf{53.6} & \textbf{67.2} \\ \hline
    ImVoxelNet & \cmark & \xmark & 28.5 & 84.4 & 73.1 & 70.1 & 51.9 & 32.2 & 15.0 & 34.2 & \phantom{0}1.6 & 29.7 & 66.1 & 23.5 & \textbf{57.8} & 43.2 & 92.4 & 54.1 & 74.0 & 34.9 & 48.1 \\ \hline
    \end{tabular} \endgroup
    \caption{AP@0.25 scores for 18 object categories from the ScanNet dataset. All methods but ImVoxelNet accept point cloud (PC) as an input.}
    \label{tab:scannet}
\end{table*}

\subsection{Results}

First, we report the results of detecting cars on outdoor KITTI and nuScenes benchmarks. Then, we discuss the results of multi-class 3D object detection on SUN RGB-D and ScanNet indoor datasets. 

\textbf{KITTI.} We present the results of monocular \textit{car} detection on KITTI in Tab. \ref{tab:kitti}. ImVoxelNet achieves the best \textit{moderate} AP on the \textit{test} split, which is the main metric in the KITTI benchmark. Moreover, our method surpasses previous state-of-the-art by 6\% AP\textsubscript{3D} and 4\% AP\textsubscript{BEV} for \textit{easy} objects. Overall, ImVoxelNet is superior in terms of almost all metrics on both \textit{test} and \textit{val} splits.

\begin{figure}[!h]
\centering
\setlength{\tabcolsep}{2pt}
\begin{tabular}{c}
    \includegraphics[width=0.9\linewidth]{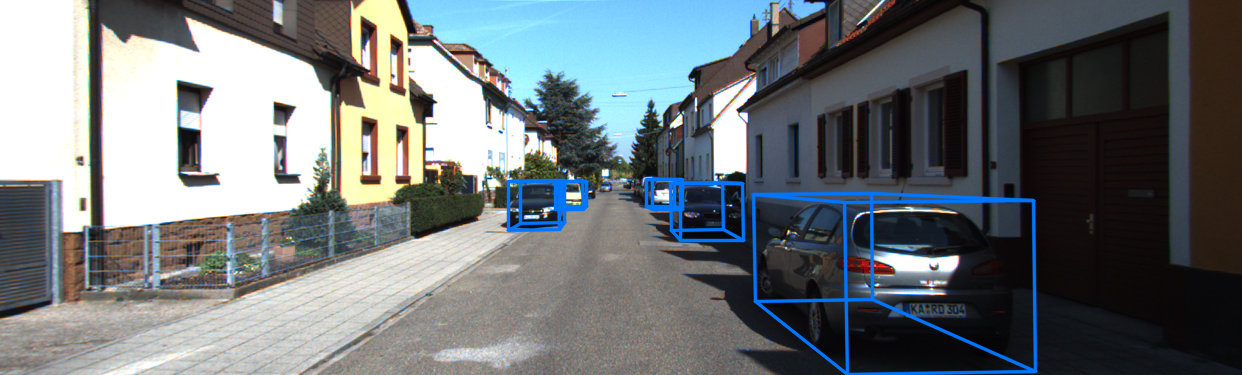} \\ \includegraphics[width=0.9\linewidth]{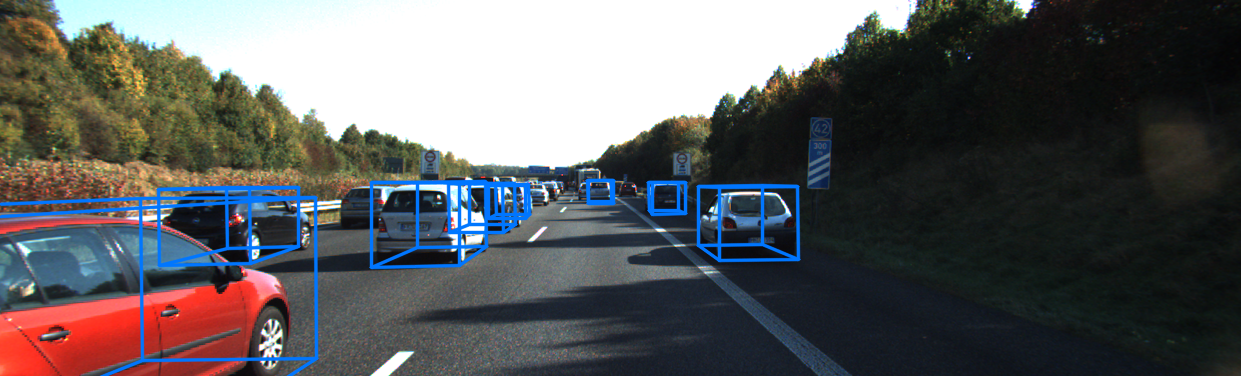}
\end{tabular}
\caption{Visualization of object detection results for monocular images from validation subset of the KITTI dataset.}
\label{fig:visualization_kitti}
\end{figure}

\textbf{nuScenes.} For nuScenes, unlike other methods that only run inference on images from 6 onboard cameras, ImVoxelNet uses multi-view inputs for training. As shown in Tab. \ref{tab:nuscenes}, the proposed method outperforms MonoDIS \cite{simonelli2020disentangling} by more than 1\% of mean AP, which is the main metric. According to AP@0.5, ImVoxelNet outputs almost twice as many highly accurate estimates comparing to MonoDIS. For car detection, two boxes might have IoU = 0 when a center distance exceeds 1 meter. By that, AP@1.0m, AP@2.0m, and AP@4.0m might be calculated for non-intersecting bounding boxes, which seems counter-intuitive (e.g., for the KITTI dataset, only boxes with IoU \textgreater 0.7 are considered to be true positive). Hence, we argue that AP@0.5 is the most decisive metric.

Moreover, we report values of ATE, ASE, and AOE metrics. As represented in the Tab. \ref{tab:nuscenes}, ImVoxelNet has at least 0.09 meters smaller ATE than other monocular methods.

\begin{figure}[!h]
\centering
\setlength{\tabcolsep}{2pt}
\begin{tabular}{cc}
    \includegraphics[width=0.45\linewidth]{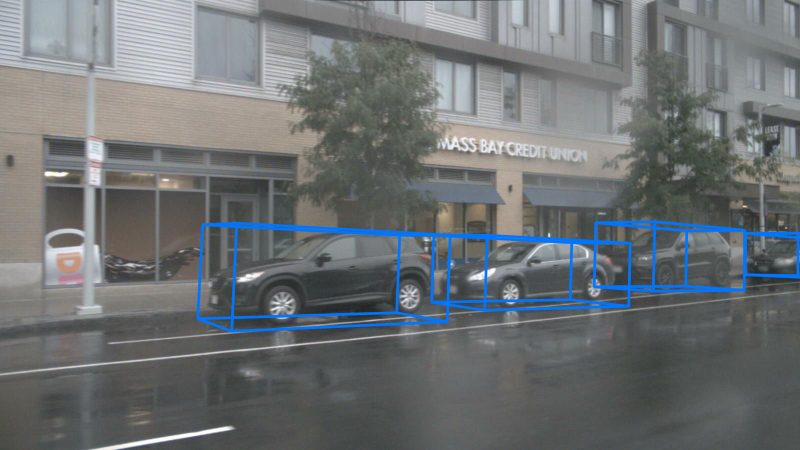} &
    \includegraphics[width=0.45\linewidth]{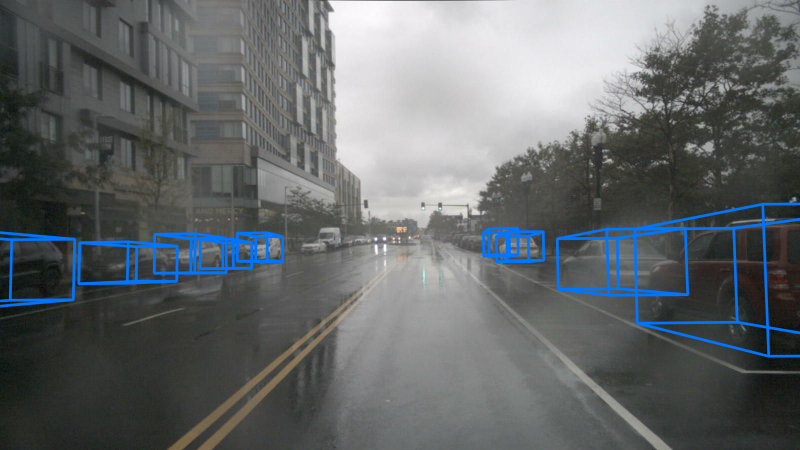} \\
    \includegraphics[width=0.45\linewidth]{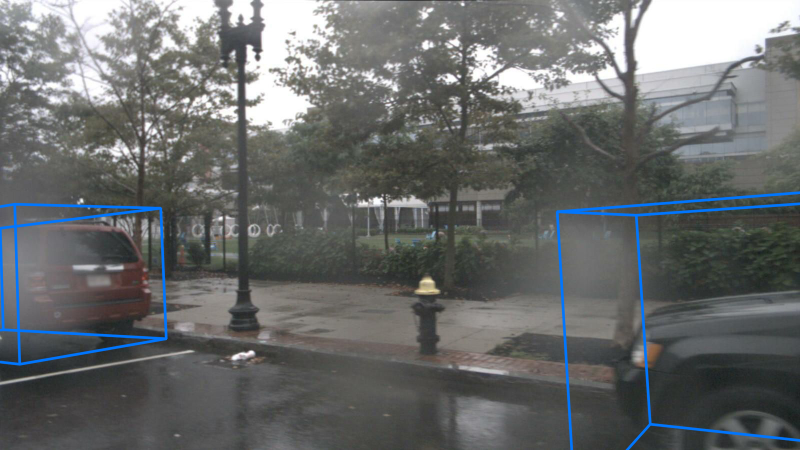} &
    \includegraphics[width=0.45\linewidth]{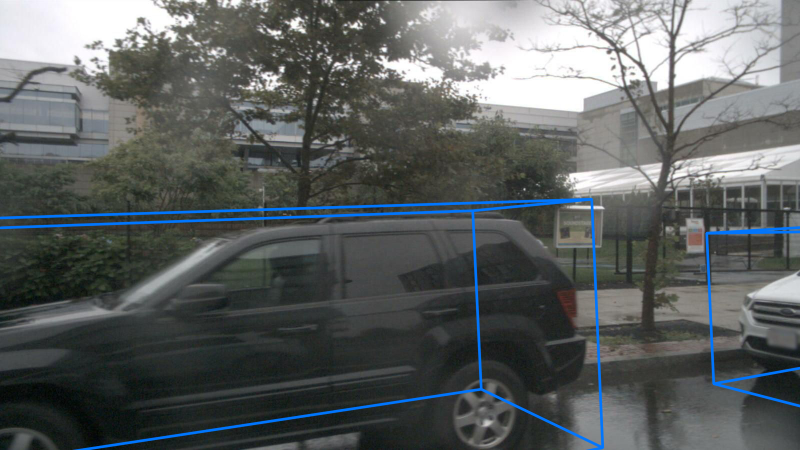}
\end{tabular}
\caption{Visualization of object detection results for multi-view inputs from validation subset (scene \textit{n008-2018-09-18-15-12-01-0400\_\_15372981046}) of the nuScenes dataset.}
\label{fig:visualization_nuscenes}
\end{figure}

\textbf{SUN RGB-D.} We compare ImVoxelNet with existing methods on the most recent monocular benchmark introduced in \cite{nie2020total3dunderstanding}, which includes objects of NYU-37 categories \cite{silberman2012nyu}. Since the chosen benchmark implies estimating camera pose and layout, we optimize $L_\mathsf{indoor} + L_\mathsf{extra}$ for training. For a fair comparison with Total3DUnderstanding \cite{nie2020total3dunderstanding}, we report their results without joint training since it requires the additional mesh-annotated dataset. Tab. \ref{tab:sunrgbd_total_10} demonstrates that ImVoxelNet surpasses all previous methods by a margin exceeding 18\% in terms of mAP. Furthermore, ImVoxelNet outperforms Total3DUnderstanding in both layout and camera pose estimation. We also report metrics on other benchmarks: the PerspectiveNet \cite{huang2019perspectivenet} benchmark with 30 object categories, and the VoteNet \cite{qi2019votenet} benchmark with 10 categories, which is used by point cloud-based methods (see \ref{sec:sunrgbd}).

\begin{figure}[!h]
\centering
\setlength{\tabcolsep}{2pt}
\begin{tabular}{cc}
    \includegraphics[width=0.45\linewidth]{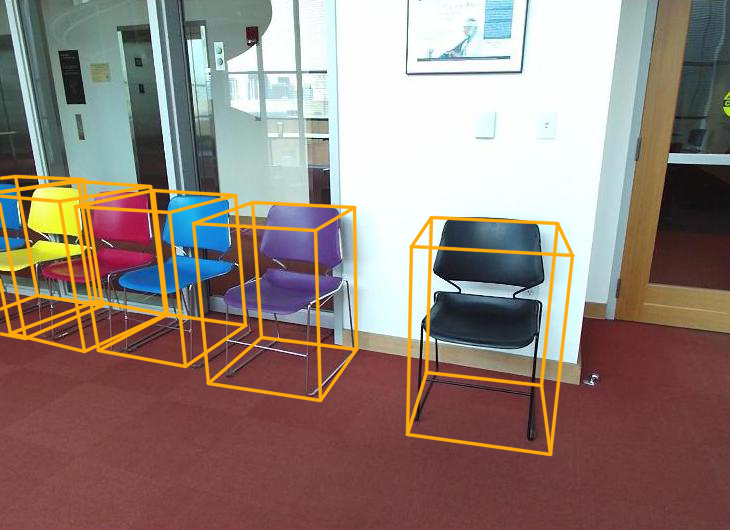} &
    \includegraphics[width=0.45\linewidth]{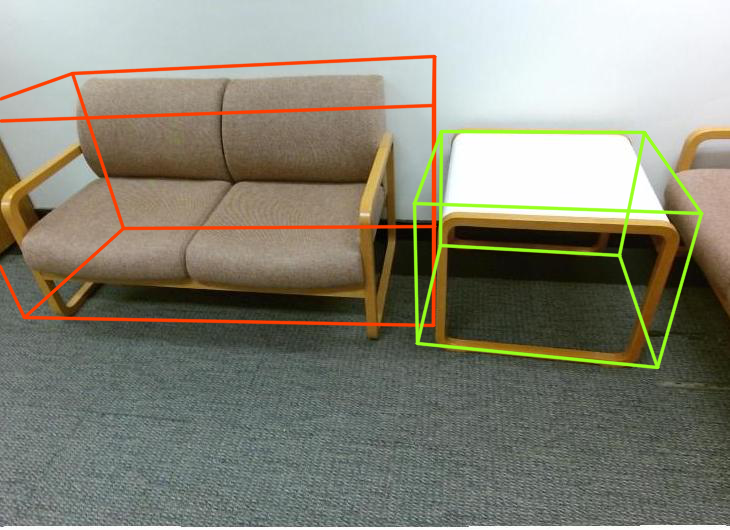} \\
    \includegraphics[width=0.45\linewidth]{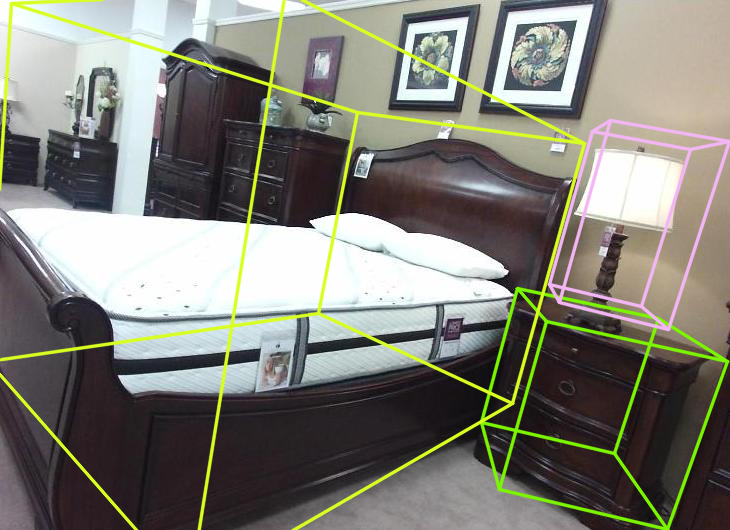} &
    \includegraphics[width=0.45\linewidth]{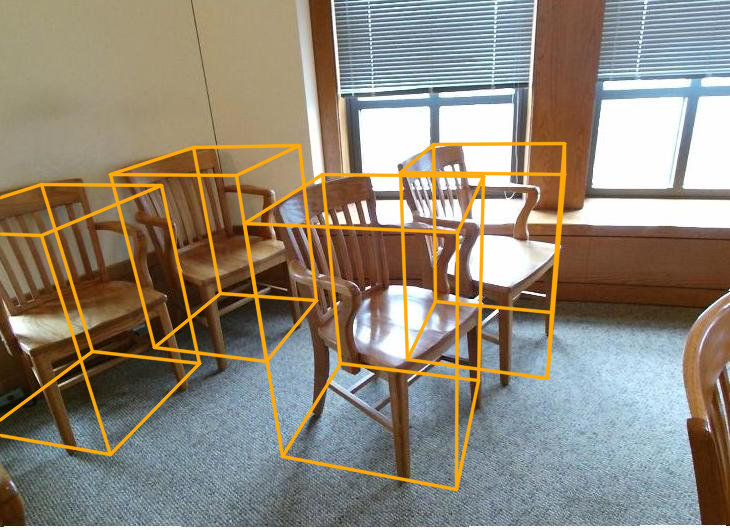}
\end{tabular}
\caption{Visualization of object detection results for monocular images from validation subset of the SUN RGB-D dataset.}
\label{fig:visualization_sun_rgbd}
\end{figure}

\textbf{ScanNet.} We compare ImVoxelNet to existing methods on the common benchmark with 18 classes. During training, we use $T=50$ images per scene, as was proposed in \cite{murez2020atlas}. We conduct an ablation study to choose an optimal number of test images per scene (Tab. \ref{tab:scannet_ablation}). We run our method five times on different samples for each number of test images and report an average result with a 0.95 confidence interval. Experiments show that the more images per test scene, the better. The most time-consuming part of the pipeline is processing a voxel volume with 3D convolutions while extracting 2D features gives a minor overhead. Consequently, with an increase in the number of test images per scene, the runtime grows sublinearly.

\begin{table}[!h]
    \centering \small
        \begingroup \setlength{\tabcolsep}{2pt}
        \begin{tabular}{l|c|c}
        \hline
        Images & mAP & Runtime[s]\\ \hline
        1 & \phantom{0}9.1 \scriptsize{$\pm$1.0} & 0.14 \\
        5 & 27.6 \scriptsize{$\pm$2.4} & 0.23 \\
        10 & 36.9 \scriptsize{$\pm$0.5} & 0.30 \\
        50 & 46.6 \scriptsize{$\pm$0.5} & 1.21 \\
        100 & \textbf{47.6} \scriptsize{$\pm$0.8} & 2.45 \\ \hline
        \end{tabular} \endgroup
        \caption{mAP@0.25 scores and runtime measured in seconds per scene for different number of images per test scene from the ScanNet dataset.}
        \label{tab:scannet_ablation}
\end{table}

According to Tab. \ref{tab:scannet}, ImVoxelNet still shows competitive results despite not using point clouds. Notably, it outperforms point cloud-based 3D-SIS \cite{hou20193dsis} which builds a voxel volume representation using RGB images as an additional modality.

\begin{figure}[!h]
\centering
\setlength{\tabcolsep}{2pt}
\begin{tabular}{ccc}
    \includegraphics[width=0.31\linewidth]{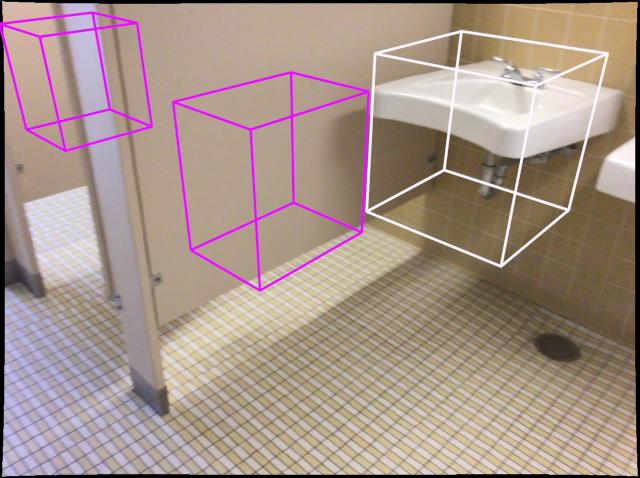} &
    \includegraphics[width=0.31\linewidth]{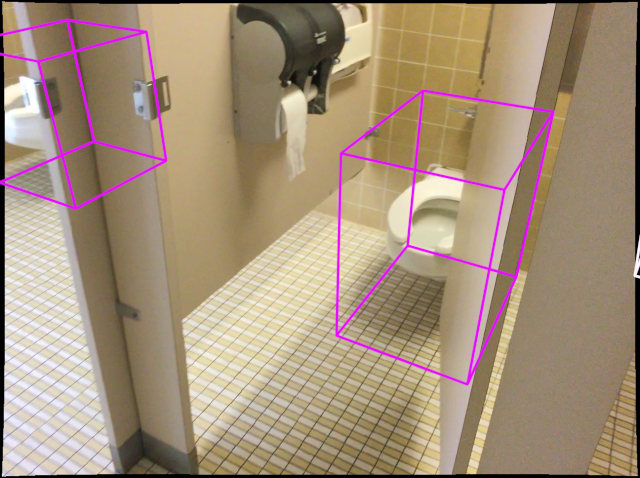} &
    \includegraphics[width=0.31\linewidth]{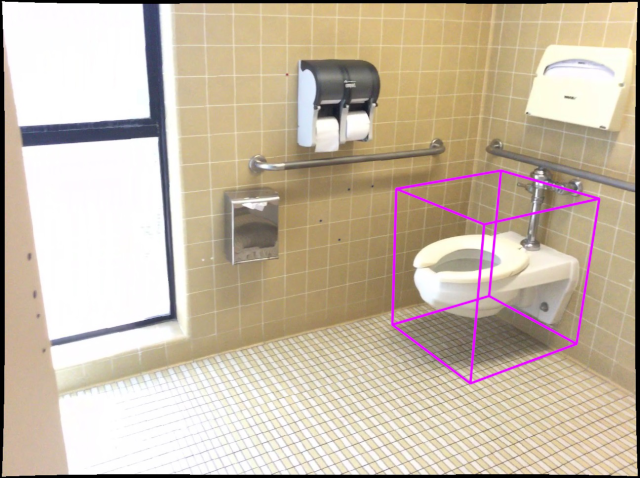}
\end{tabular}
\caption{Visualization of object detection results for multi-view inputs from validation subset (scene \textit{0086\_00}) of the ScanNet dataset.}
\label{fig:visualization_scannet}
\end{figure}

\textbf{Performance.} We report the inference time on the KITTI dataset in Tab. \ref{tab:performance}. All the methods were examined in the same experimental setup on a single GPU. ImVoxelNet uses computationally expensive 3D convolutions, so it is expected to be slower than the methods that rely on 2D convolutions only. In our experiments, ImVoxelNet appeared to be inferior in speed to most of the listed methods, yet the runtime differs within an order of magnitude. The listed methods use different backbones, and this affects the total speed. In ImVoxelNet, extracting features with a backbone is a simple, lightweight procedure compared to processing voxel volume with 3D convolutions. Accordingly, the choice of a backbone is negligible: experiments show that replacing ResNet-50 with a more lightweight version has a minor influence on performance.

\begin{table}[!h]
\centering \small
    \begingroup \setlength{\tabcolsep}{2pt}
    \begin{tabular}{l|c|c|c}
    \hline
    Method & Backbone & AP & Runtime[s] \\ \hline
    OFTNet\cite{roddick2018orthographic} & ResNet-18 & \phantom{0}3.27 & 0.50 \\
    GS3D\cite{li2019gs3d} & VGG-16 & 10.97 & 2.00 \\
    MonoGRNet\cite{qin2019monogrnet} & VGG-16 & 10.19 & 0.06 \\
    MonoDIS\cite{simonelli2020disentangling} & ResNet-34 & 14.98 & 0.10 \\
    SMOKE\cite{liu2020smoke} & DLA-34 & 12.85 & \textbf{0.03} \\
    M3D-RPN\cite{brazil2019m3drpn} & DenseNet-121 & 17.06 & 0.16 \\
    \multirow[l]{3}{*}{ImVoxelNet} & ResNet-18 & 16.23 & 0.37 \\
    & ResNet-34 & 16.58 & 0.38 \\
    & ResNet-50 & \textbf{17.80} & 0.40 \\ \hline
    \end{tabular} \endgroup
    \caption{AP\textsubscript{3D}@0.7 for \textit{car} category, \textit{moderate} difficulty and runtime measured in seconds per image, estimated for the validation subset of the KITTI dataset.}
\label{tab:performance}
\end{table}

\section{Conclusion}

In this paper, we formulate the task of multi-view RGB-based 3D object detection as an end-to-end optimization problem. To address this problem, we have proposed ImVoxelNet, a novel fully convolutional method of 3D object detection given posed monocular or multi-view RGB inputs. During both training and inference, ImVoxelNet accepts multi-view inputs with an arbitrary number of views. Besides, our method can accept monocular inputs (treated as a special case of multi-view inputs). The proposed method has achieved state-of-the-art results in outdoor car detection on both the monocular KITTI benchmark and the multi-view nuScenes benchmark. Moreover, it has surpassed existing methods of 3D object detection on the indoor SUN RGB-D dataset. For the ScanNet dataset, ImVoxelNet has set a new benchmark for indoor multi-view 3D object detection. Overall, ImVoxelNet successfully works on both indoor and outdoor data, which makes it general-purpose.

\clearpage

\begingroup
    \small
    \bibliographystyle{abbrvnat}
    \bibliography{main}
\endgroup

\clearpage

\section*{Appendix}

\appendix

\section{More results on SUN RGB-D} \label{sec:sunrgbd}

For a comprehensive comparison, we also mention PerspectiveNet \cite{huang2019perspectivenet}, which is evaluated following a different protocol. In that protocol, the annotations are mapped into 30 object categories. Accordingly, we train ImVoxelNet using the same object categories. The results are reported in Tab. \ref{tab:sunrgbd_perspective_30}. Among these 30 categories, 10 object categories are consistent with 10 categories used in \cite{huang2018holistic, huang2018cooperative, nie2020total3dunderstanding}. So, we can merge these benchmarks and report metrics for \cite{huang2018holistic, huang2018cooperative, nie2020total3dunderstanding, huang2019perspectivenet} that are obtained on the same subset of 10 object categories \ref{tab:sunrgbd_perspective_10}. Following \cite{huang2019perspectivenet}, we assume camera poses are known, so we optimize only $L_{indoor}$ and do not use any additional camera pose loss.

Another SUN RGB-D benchmark has been proposed in \cite{qi2019votenet} for point cloud-based methods evaluation. This benchmark implies detecting objects of 10 categories with mAP@0.25 chosen as the main metric. In Tab. \ref{tab:sunrgbd25}, we report the results of our method against point cloud-based methods. This comparison is unfair, favoring point cloud-based methods since they have access to more complete data. Nevertheless, we report the metrics to establish a baseline for monocular 3D object detection on SUN RGB-D.

Comparison with Total3DUnderstanding \cite{nie2020total3dunderstanding} on all NYU-37 object categories is present in Tab. \ref{tab:sunrgbd_total_37}. In this experiment, we optimize $L_\mathsf{indoor}+L_\mathsf{extra}$ since camera pose is assumed unknown.

\section{Visualization} \label{sec:visualization}

All visualized images belong to validation subsets of the corresponding datasets. Different colors of the depicted bounding boxes mark different object categories; the color encoding is consistent within each dataset.

\begin{table*}[!ht]
    \centering \small
    \begingroup \setlength{\tabcolsep}{2pt}
    \begin{tabular}{l|cc|cccccccccc|c}
    \hline
    Method & RGB & PC & bath & bed & bkshf & chair & desk & dresser & nstand & sofa & table & toilet & mAP \\ \hline
    F-PointNet\cite{qi2018frustum} & \cmark & \cmark & 43.3 & 81.1 & 33.3 & 64.2 & 24.7 & 32.0 & 58.1 & 61.1 & 51.1 & \textbf{90.9} & 54.0 \\
    VoteNet\cite{qi2019votenet} & \xmark & \cmark & 74.4 & 83.0 & 28.8 & 75.3 & 22.0 & 29.8 & 62.2 & 64.0 & 47.3 & 90.1 & 57.7 \\
    H3DNet\cite{zhang2020h3dnet} & \xmark & \cmark & 73.8 & 85.6 & 31.0 & \textbf{76.7} & \textbf{29.6} & 33.4 & 65.5 & 66.5 & 50.8 & 88.2 & 60.1 \\
    ImVoteNet\cite{qi2020imvotenet} & \cmark & \cmark & \textbf{75.9} & \textbf{87.6} & \textbf{41.3} & \textbf{76.7} & 28.7 & \textbf{41.4} & \textbf{69.9} & \textbf{70.7} & \textbf{51.1} & 90.5 & \textbf{63.4} \\ \hline
    ImVoxelNet & \cmark & \xmark & 71.7 & 69.6 & \phantom{0}5.7 & 53.7 & 21.9 & 21.2 & 34.6 & 51.5 & 39.1 & 76.8 & 40.7 \\ \hline
    \end{tabular} \endgroup
    \caption{AP@0.25 scores for 10 object categories \cite{qi2019votenet} from the SUN RGB-D dataset. All methods but ImVoxelNet use point cloud (PC) as an input.}
    \label{tab:sunrgbd25}
\end{table*}

\begin{table*}[!ht]
    \centering \small
    \begingroup \setlength{\tabcolsep}{2pt}
    \begin{tabular}{l|cccccccccc|c}
    \hline
    Method & bed & chair & sofa & table & desk & toilet & bin & sink & shelf & lamp & mAP \\ \hline
    3DGP\cite{choi20133dgp} & \phantom{0}5.62 & \phantom{0}2.31 & \phantom{0}3.24 & \phantom{0}1.23 & -- & -- & -- & -- & -- & -- & -- \\
    HoPR\cite{huang2018holistic} & 58.29 & 13.56 & 28.37 & 12.12 & \phantom{0}4.79 & 16.50 & \phantom{0}0.63 & \phantom{0}2.18 & \phantom{0}1.29 & \phantom{0}2.41 & 14.01 \\
    CooP\cite{huang2018cooperative} & 63.58 & 17.12 & 41.22 & 26.21 & \phantom{0}9.55 & 58.55 & 10.19 & \phantom{0}5.34 & \phantom{0}3.01 & \phantom{0}1.75 & 23.65 \\
    PerspectiveNet\cite{huang2019perspectivenet} & \textbf{79.69} & 40.42 & 62.35 & 44.12 & 20.19 & 81.22 & 22.42 & \textbf{41.35} & 8.29 & 13.14 & 39.09 \\
    ImVoxelNet & 77.87 & \textbf{65.94} & \textbf{63.89} & \textbf{51.17} & \textbf{31.91} & \textbf{84.53} & \textbf{33.35} & 39.91 & \textbf{21.65} & \textbf{17.19} & \textbf{48.74} \\ \hline
    \end{tabular} \endgroup
    \caption{AP@0.15 scores for 10 out of 30 object categories \cite{huang2019perspectivenet} from the SUN RGB-D dataset.}
    \label{tab:sunrgbd_perspective_10}
\end{table*}

\begin{table*}[!ht]
    \newcolumntype{A}{>{\centering\arraybackslash}p{0.063\textwidth}}
    \centering \small
    \begingroup \setlength{\tabcolsep}{2pt} \begin{align*}
    & \begin{tabular}{l|AAAAAAAAAAA}
        \hline
        Method & toilet & recycle bin & night stand & end table & drawer & computer & key board & table & chair & monitor & stool \\ \hline
        PerspectiveNet\cite{huang2019perspectivenet} & 81.22 & 37.68 & 35.16 & 19.77 & 1.28 & 1.24 & \textbf{2.86} & 44.12 & 40.42 & 1.14 & \textbf{22.65} \\
        ImVoxelNet & \textbf{84.53} & \textbf{52.20} & \textbf{46.29} & \textbf{25.31} & \textbf{6.05} & \textbf{2.71} & 0.01 & \textbf{51.17} & \textbf{65.94} & \textbf{19.82} & 10.37 \\ \hline
    \end{tabular} \vspace{\baselineskip} \\
    & \begin{tabular}{l|AAAAAAAAAAA}
        \hline
        Method & lamp & dresser & picture & garbage bin & shelf & sofa chair & cabinet & sink & desk & book shelf & coffee table \\ \hline
        PerspectiveNet\cite{huang2019perspectivenet} & 13.14 & \textbf{27.38} & 0.00 & 22.42 & 0.97 & 51.86 & 1.70 & \textbf{41.35} & 20.19 & 8.29 & 28.80 \\
        ImVoxelNet & \textbf{17.19} & 22.32 & \textbf{0.82} & \textbf{33.35} & \textbf{4.00} & \textbf{54.61} & \textbf{7.90} & 39.91 & \textbf{31.91} & \textbf{21.65} & \textbf{36.48} \\ \hline
    \end{tabular} \vspace{\baselineskip} \\
    & \begin{tabular}{l|AAAAAAAA}
        \hline
        Method & box & sofa & white board & bed & pillow & paper & painting & cpu \\ \hline
        PerspectiveNet\cite{huang2019perspectivenet} & 1.64 & 62.35 & 0.02 & \textbf{79.69} & 11.36 & 0.00 & 0.17 & \textbf{21.60} \\
        ImVoxelNet & \textbf{3.29} & \textbf{63.89} & \textbf{0.95} & 77.87 & \textbf{14.65} & 0.00 & \textbf{0.53} & 5.30 \\ \hline
    \end{tabular}
    \end{align*} \endgroup
    \caption{AP@0.15 scores for 30 object categories \cite{huang2019perspectivenet} from the SUN RGB-D dataset.}
    \label{tab:sunrgbd_perspective_30}
\end{table*}

\begin{table*}[!ht]
    \newcolumntype{A}{>{\centering\arraybackslash}p{0.063\textwidth}}
    \centering \small
    \begingroup \setlength{\tabcolsep}{2pt} \begin{align*}
    & \begin{tabular}{l|AAAAAAAAAAA}
        \hline
        Method & cabinet & bed & chair & sofa & table & door & window & book shelf & picture & counter & blinds \\ \hline
        CooP\cite{huang2018cooperative} & 10.47 & 57.71 & 15.21 & 36.67 & 31.16 & 0.14 & 0.00 & 3.81 & 0.00 & 27.67 & \textbf{2.27} \\
        T3DU\cite{nie2020total3dunderstanding} & 11.39 & 59.03 & 15.98 & 43.95 & 35.28 & 0.36 & 0.16 & 5.26 & \textbf{0.24} & \textbf{33.51} & 0.00 \\
        ImVoxelNet & \textbf{19.24} & \textbf{79.17} & \textbf{63.07} & \textbf{60.59} & \textbf{51.14} & \textbf{0.74} & \textbf{0.18} & \textbf{16.37} & 0.14 & 14.89 & 0.26 \\ \hline
    \end{tabular} \vspace{\baselineskip} \\
    & \begin{tabular}{l|AAAAAAAAAAA}
        \hline
        Method & desk & shelves & curtain & dresser & pillow & mirror & floor mat & clothes & books & fridge & tv \\ \hline
        CooP\cite{huang2018cooperative} & 19.90 & 2.96 & 1.35 & 15.98 & 2.53 & \textbf{0.47} & -- & 0.00 & \textbf{3.19} & 21.50 & 5.20 \\
        T3DU\cite{nie2020total3dunderstanding} & 23.65 & 4.96 & 2.68 & 19.20 & 2.99 & 0.19 & -- & 0.00 & 1.30 & 20.68 & 4.44 \\
        ImVoxelNet & \textbf{31.20} & \textbf{5.47} & \textbf{3.34} & \textbf{35.45} & \textbf{11.01} & 0.22 & -- & \textbf{1.40} & 0.13 & \textbf{23.28} & \textbf{12.41} \\ \hline
    \end{tabular} \vspace{\baselineskip} \\
    & \begin{tabular}{l|AAAAAAAAAAA}
        \hline
        Method & paper & towel & shower curtain & box & white board & person & night stand & toilet & sink & lamp & bathtub \\ \hline
        CooP\cite{huang2018cooperative} & 0.20 & 2.14 & \textbf{20.00} & 2.59 & 0.16 & 20.96 & 11.36 & 42.53 & 15.95 & 3.28 & 24.71 \\
        T3DU\cite{nie2020total3dunderstanding} & \textbf{0.41} & \textbf{2.20} & \textbf{20.00} & 2.25 & 0.43 & 23.36 & 6.87 & 48.37 & 14.40 & 3.46 & 27.85 \\
        ImVoxelNet & 0.00 & 1.92 & 0.00 & \textbf{2.71} & \textbf{1.17} & \textbf{42.02} & \textbf{38.38} & \textbf{77.28} & \textbf{45.12} & \textbf{13.27} & \textbf{43.59} \\ \hline
    \end{tabular} \vspace{\baselineskip} \\
    & \begin{tabular}{l|AAAA}
        \hline
        Method & bag & wall & floor & ceiling \\ \hline
        CooP\cite{huang2018cooperative} & 1.53 & -- & -- & -- \\
        T3DU\cite{nie2020total3dunderstanding} & \textbf{2.27} & -- & -- & -- \\
        ImVoxelNet & 0.53 & -- & -- & -- \\ \hline
    \end{tabular}
    \end{align*} \endgroup
    \caption{AP@0.15 scores for 37 object categories \cite{nie2020total3dunderstanding} from the SUN RGB-D dataset.}
    \label{tab:sunrgbd_total_37}
\end{table*}

\newpage

\begin{figure*}[!ht]
\centering
\setlength{\tabcolsep}{2pt}
\renewcommand{\arraystretch}{0.75}
\begin{tabular}{ccc}
    \includegraphics[width=0.32\linewidth]{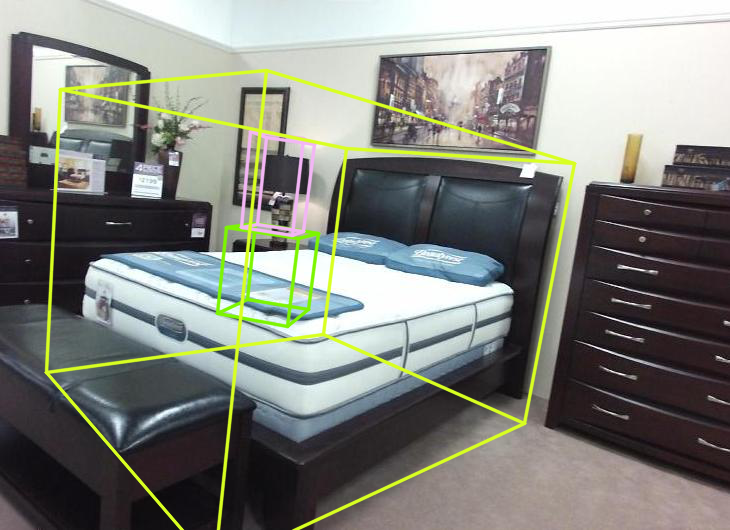} &
    \includegraphics[width=0.32\linewidth]{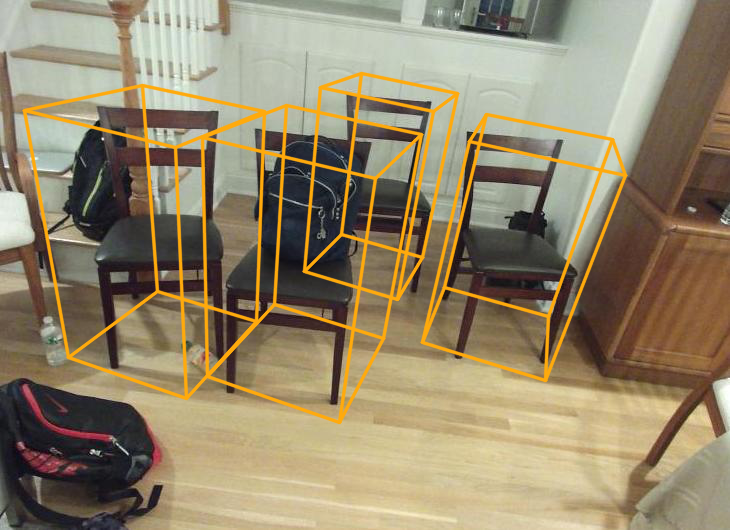} &
    \includegraphics[width=0.32\linewidth]{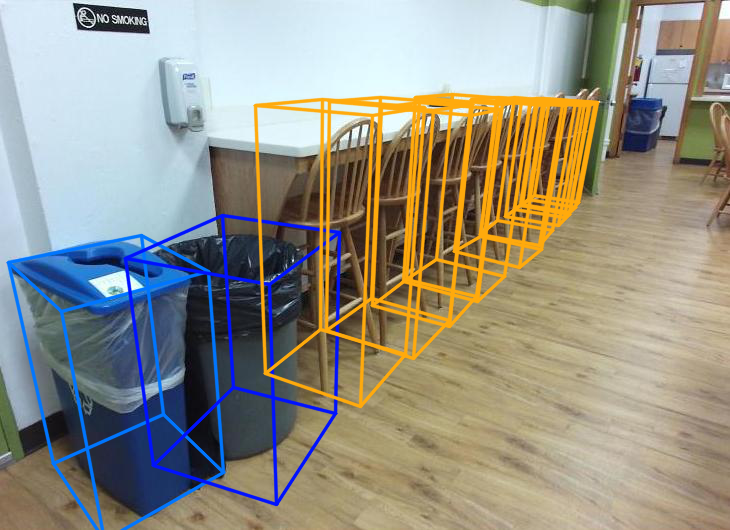} \\
\end{tabular}
\begin{tabular}{ccc}
    \includegraphics[width=0.32\linewidth]{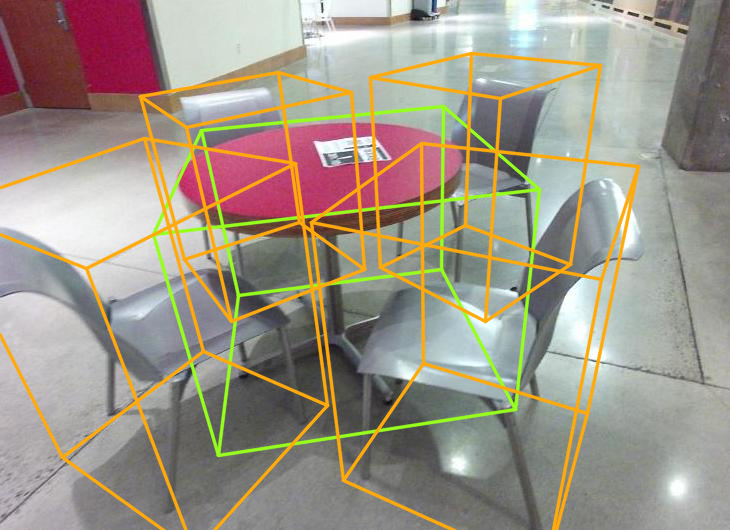} &
    \includegraphics[width=0.32\linewidth]{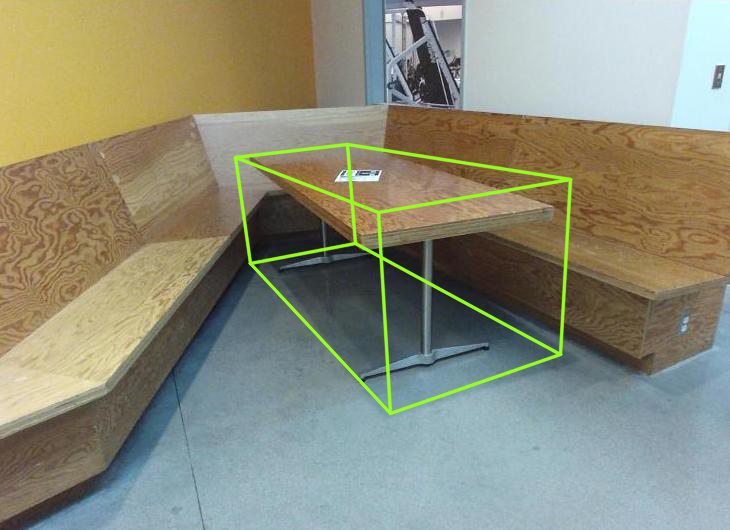} &
    \includegraphics[width=0.32\linewidth]{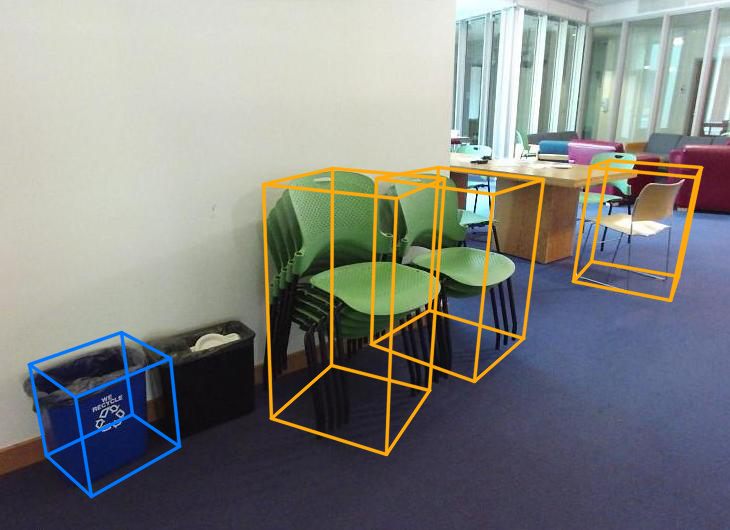} \\
\end{tabular}
\begin{tabular}{ccc}
    \includegraphics[width=0.32\linewidth]{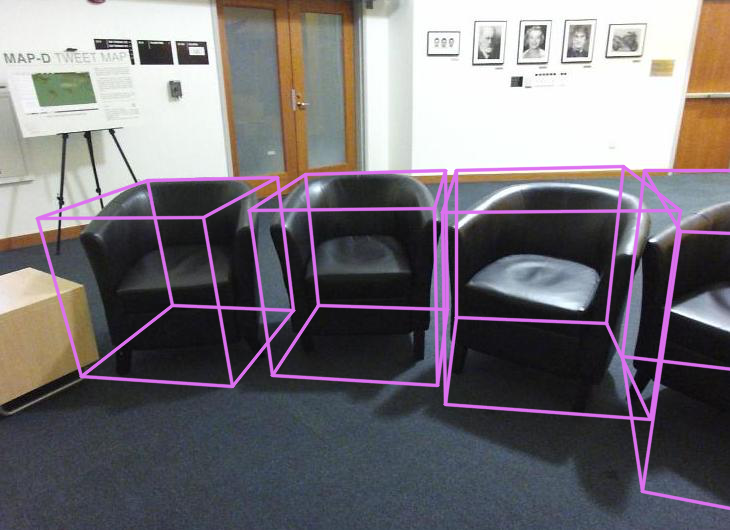} &
    \includegraphics[width=0.32\linewidth]{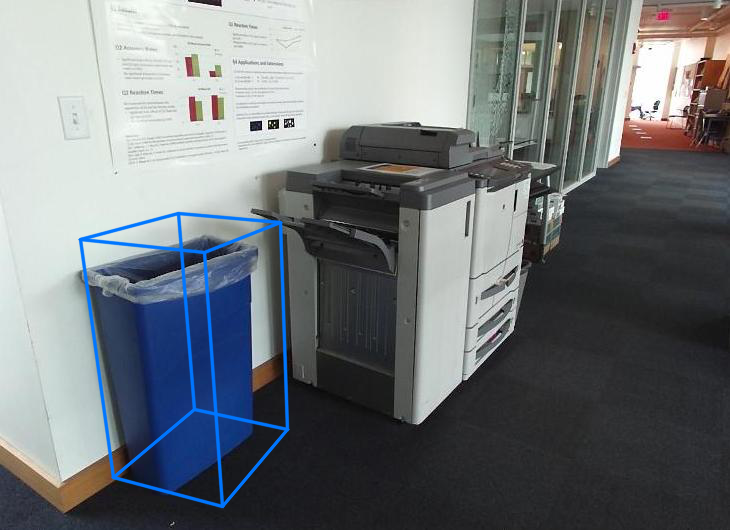} &
    \includegraphics[width=0.32\linewidth]{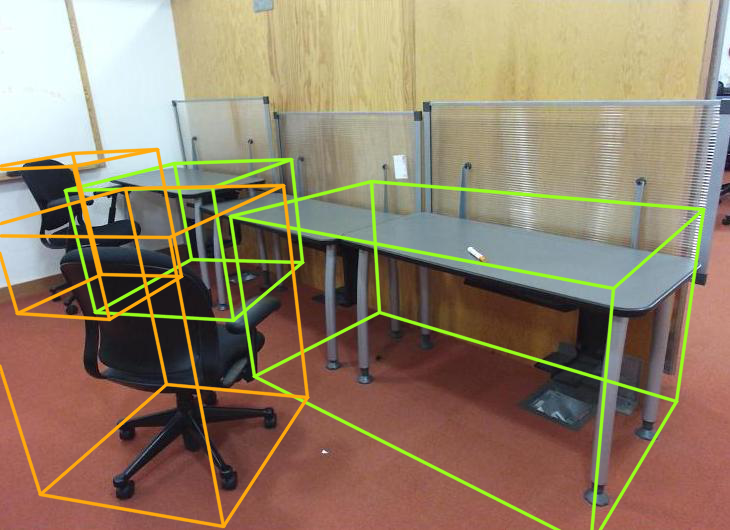} \\
\end{tabular}
\begin{tabular}{ccc}
    \includegraphics[width=0.32\linewidth]{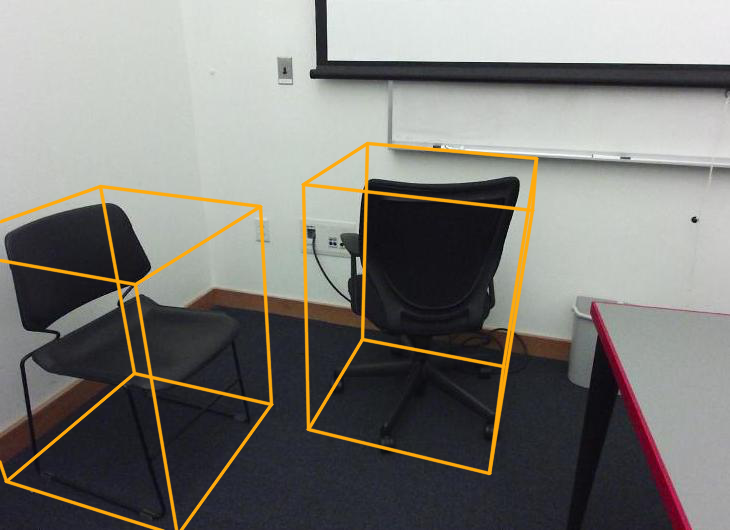} &
    \includegraphics[width=0.32\linewidth]{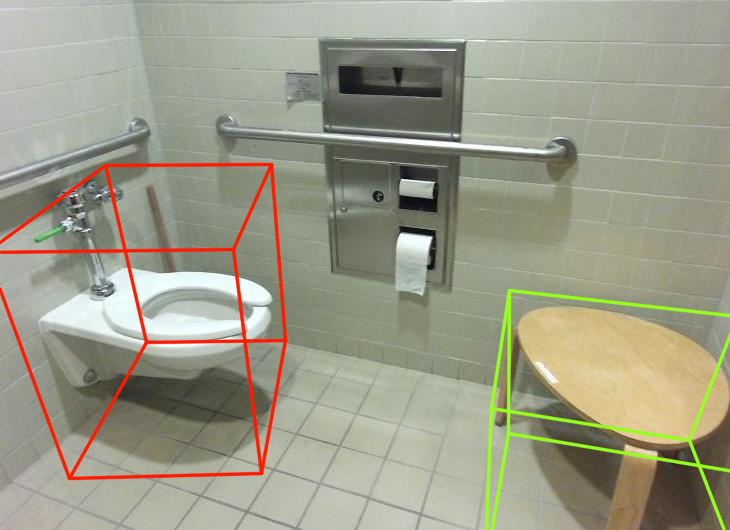} &
    \includegraphics[width=0.32\linewidth]{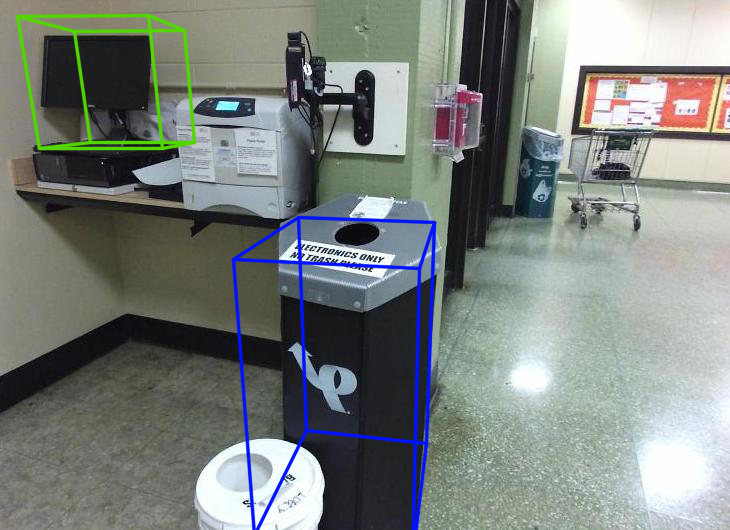} \\
\end{tabular}
\begin{tabular}{ccc}
    \includegraphics[width=0.32\linewidth]{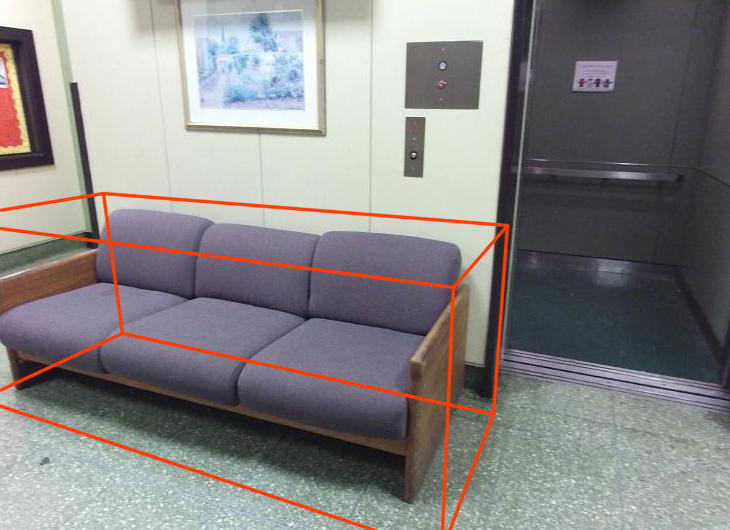} &
    \includegraphics[width=0.32\linewidth]{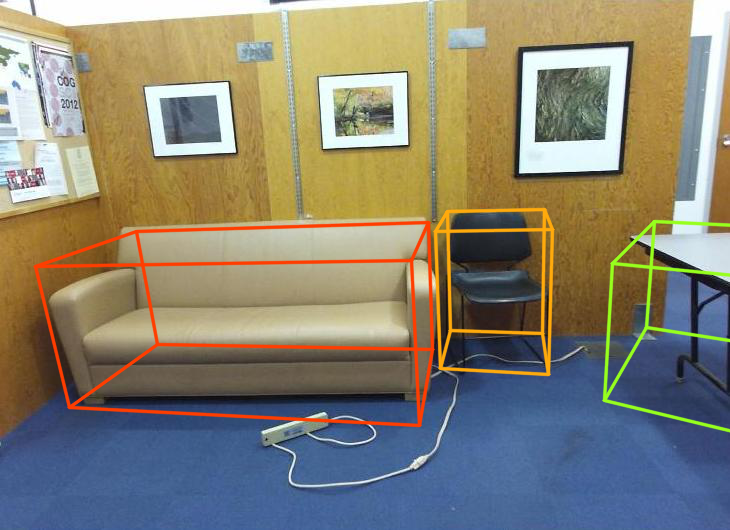} &
    \includegraphics[width=0.32\linewidth]{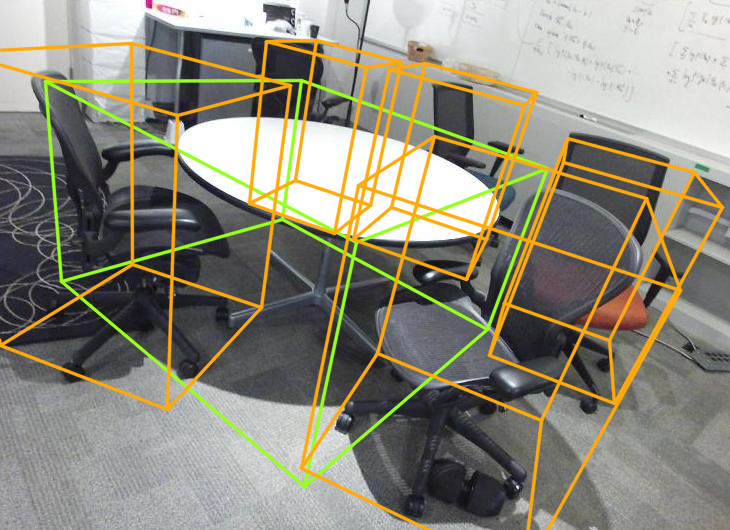} \\
\end{tabular}
\caption{Objects detected on the monocular images from the validation subset of the SUN RGB-D dataset.}
\label{fig:vis_sunrgbd}
\end{figure*}

\begin{figure*}[!ht]
\centering
\setlength{\tabcolsep}{2pt}
\renewcommand{\arraystretch}{0.75}
\begin{tabular}{ccc}
    \includegraphics[width=0.32\linewidth]{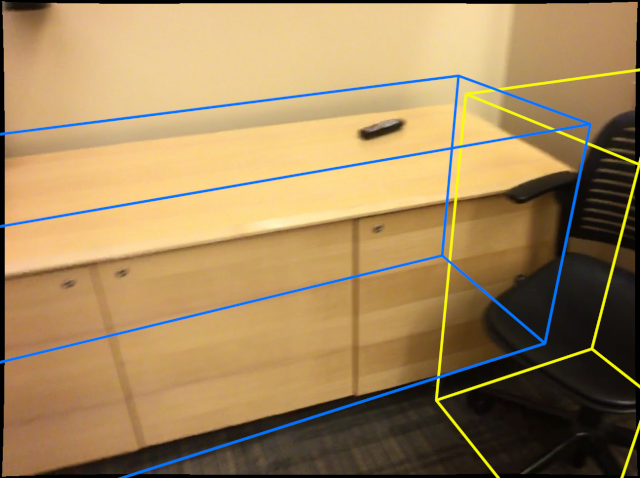} &
    \includegraphics[width=0.32\linewidth]{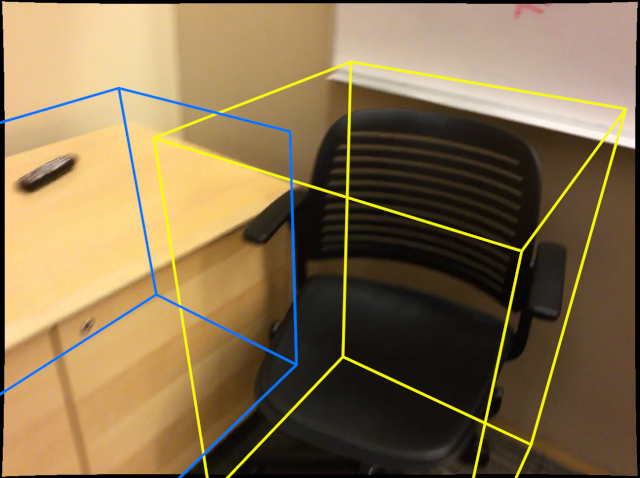} &
    \includegraphics[width=0.32\linewidth]{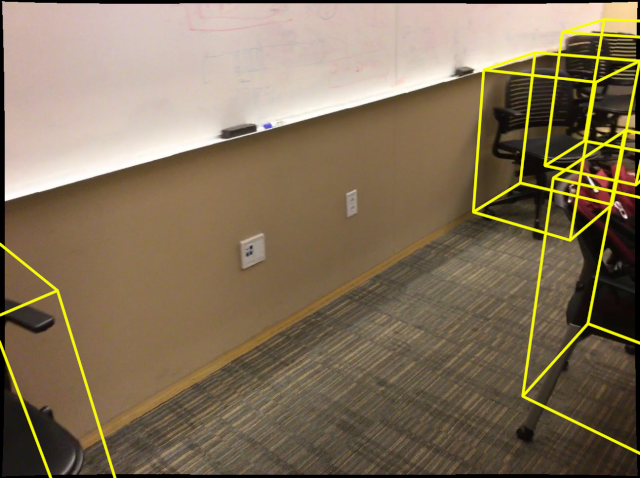} \\
\end{tabular}
\begin{tabular}{ccc}
    \includegraphics[width=0.32\linewidth]{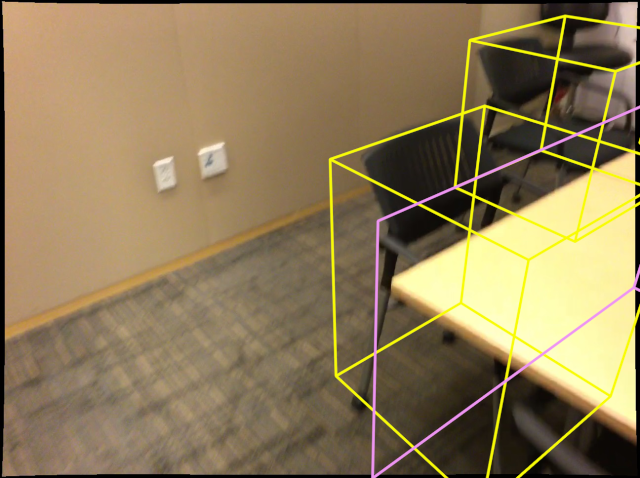} &
    \includegraphics[width=0.32\linewidth]{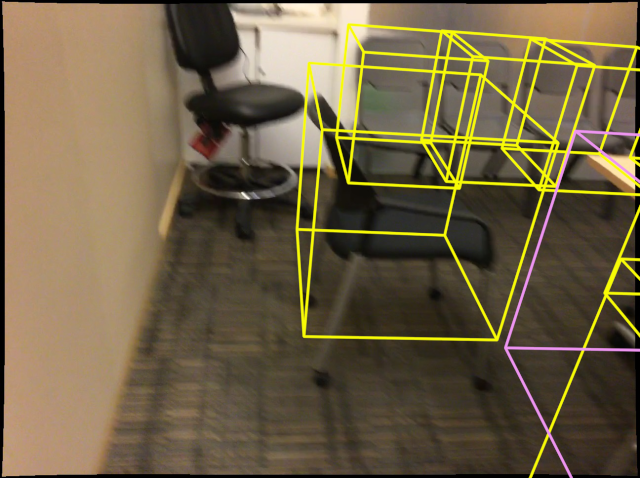} &
    \includegraphics[width=0.32\linewidth]{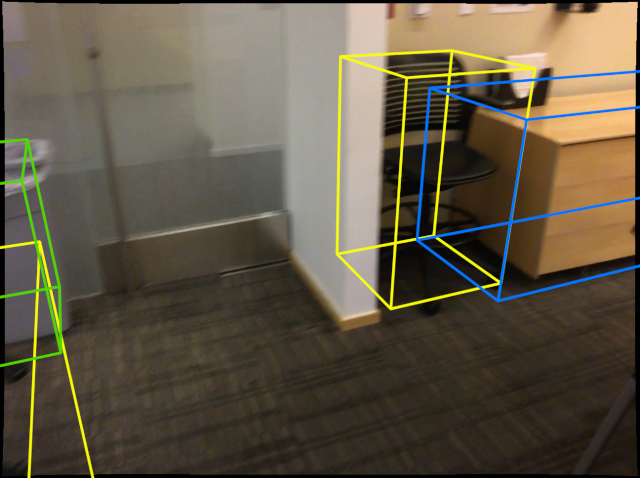} \\
\end{tabular}
a) Scene \textit{0169\_00}.
\\ \vspace{3pt}
\begin{tabular}{ccc}
    \includegraphics[width=0.32\linewidth]{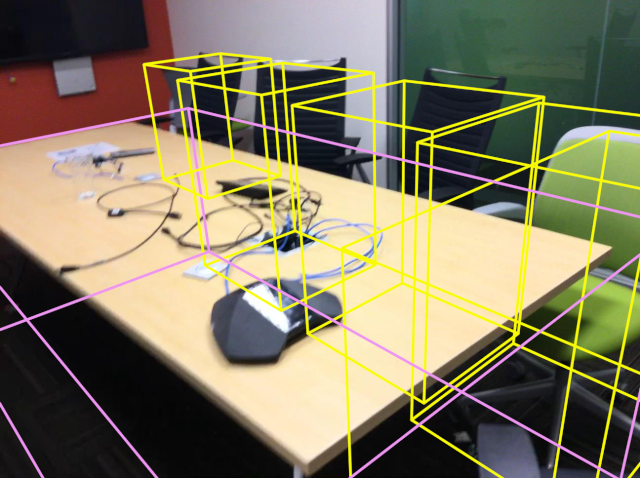} &
    \includegraphics[width=0.32\linewidth]{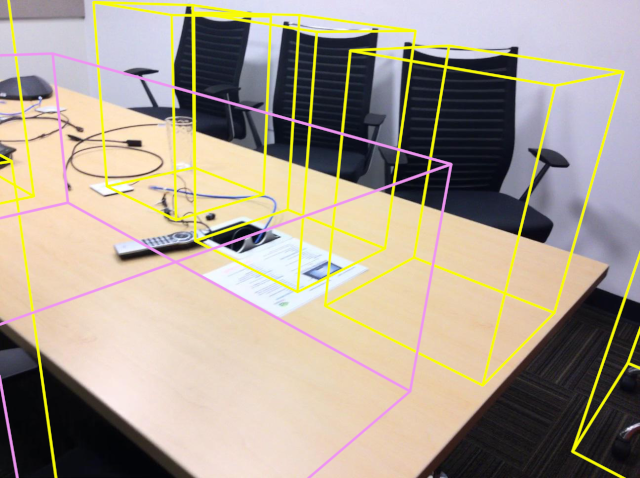} &
    \includegraphics[width=0.32\linewidth]{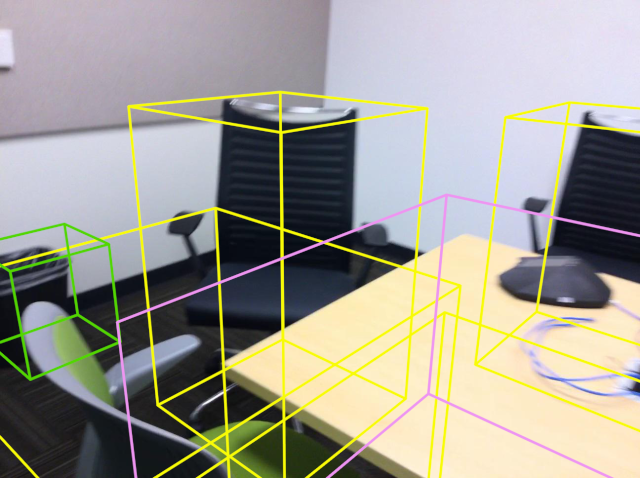} \\
\end{tabular}
\begin{tabular}{ccc}
    \includegraphics[width=0.32\linewidth]{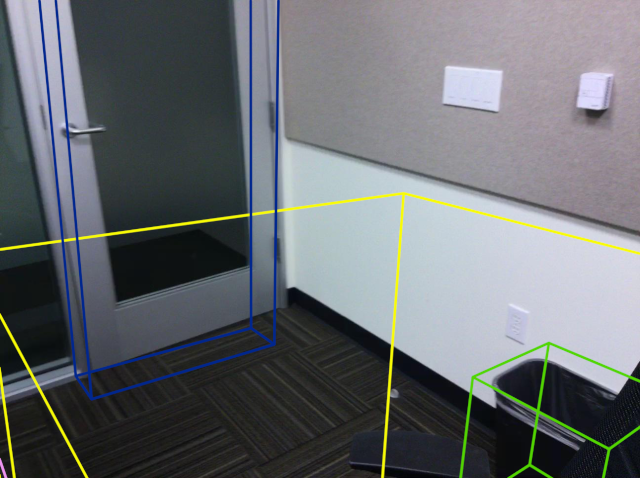} &
    \includegraphics[width=0.32\linewidth]{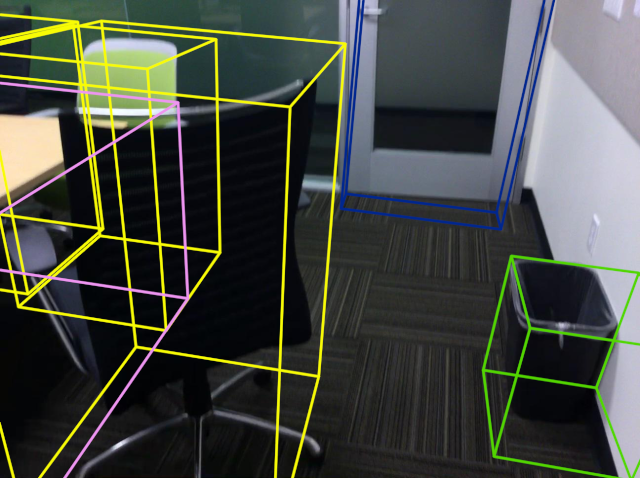} &
    \includegraphics[width=0.32\linewidth]{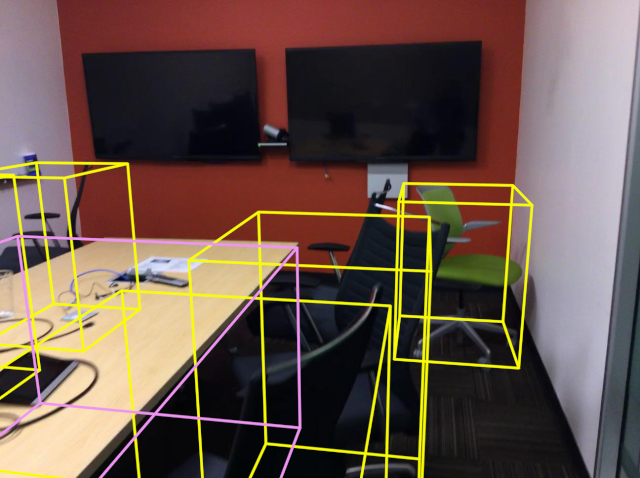} \\
\end{tabular}
b) Scene \textit{0575\_00}.
\caption{Objects detected on the multi-view inputs from the validation subset of the ScanNet dataset.}
\label{fig:vis_scannet}
\end{figure*}

\begin{figure*}[!ht]
\centering
\setlength{\tabcolsep}{2pt}
\renewcommand{\arraystretch}{0.75}
\begin{tabular}{c}
    \includegraphics[width=0.8\linewidth]{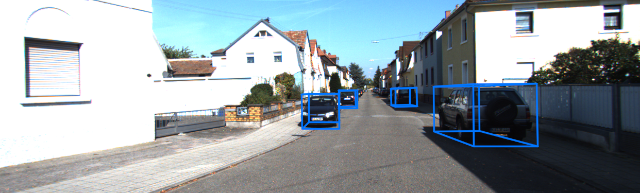}
\end{tabular}
\begin{tabular}{c}
    \includegraphics[width=0.8\linewidth]{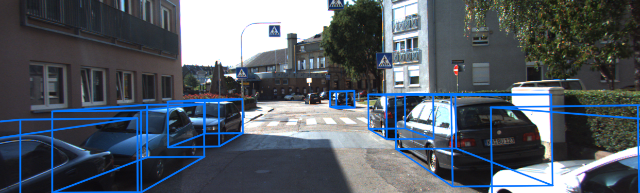}
\end{tabular}
\begin{tabular}{c}
    \includegraphics[width=0.8\linewidth]{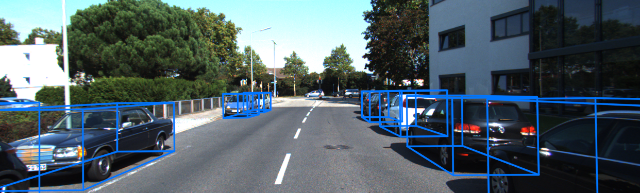}
\end{tabular}
\begin{tabular}{c}
    \includegraphics[width=0.8\linewidth]{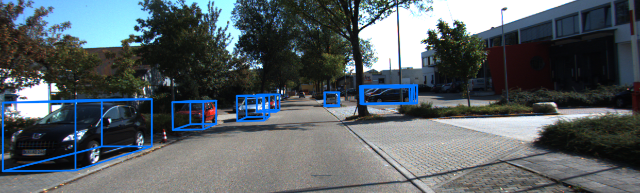}
\end{tabular}
\begin{tabular}{c}
    \includegraphics[width=0.8\linewidth]{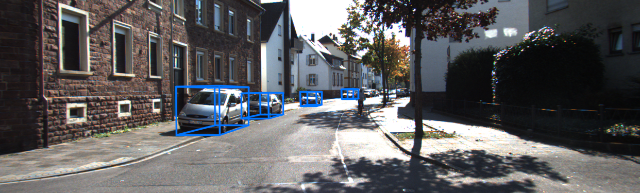}
\end{tabular}
\caption{Cars detected on the monocular images from the validation subset of the KITTI dataset.}.
\label{fig:vis_kitti}
\end{figure*}

\begin{figure*}[!ht]
\centering
\setlength{\tabcolsep}{2pt}
\renewcommand{\arraystretch}{0.75}
\begin{tabular}{ccc}
    \includegraphics[width=0.32\linewidth]{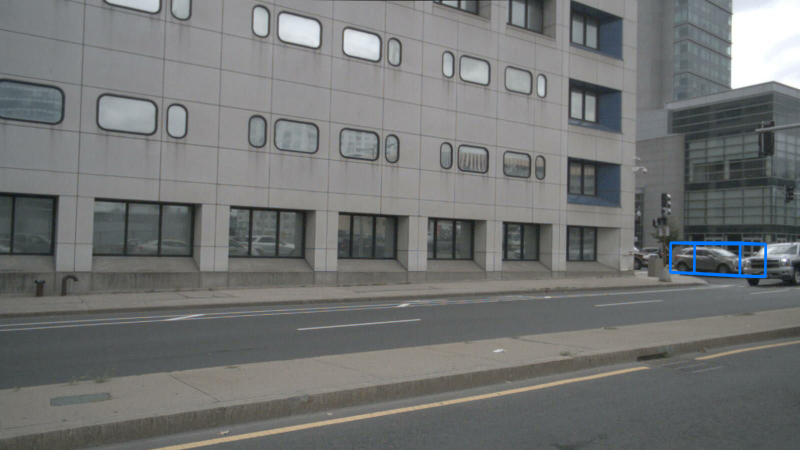} &
    \includegraphics[width=0.32\linewidth]{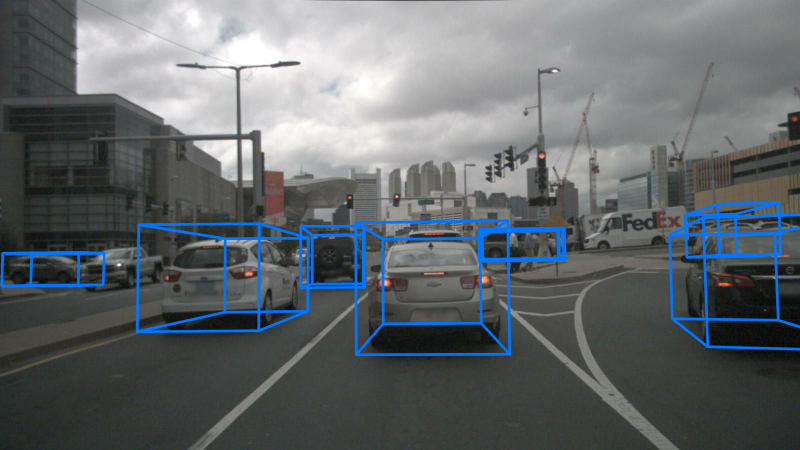} &
    \includegraphics[width=0.32\linewidth]{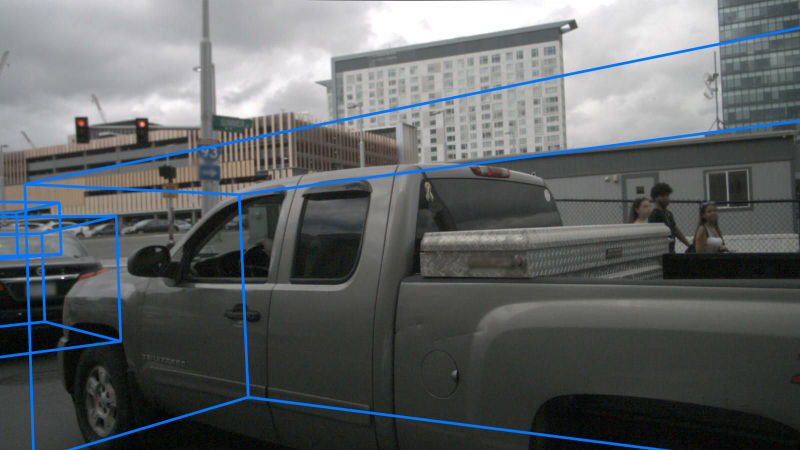} \\
\end{tabular}
\begin{tabular}{ccc}
    \includegraphics[width=0.32\linewidth]{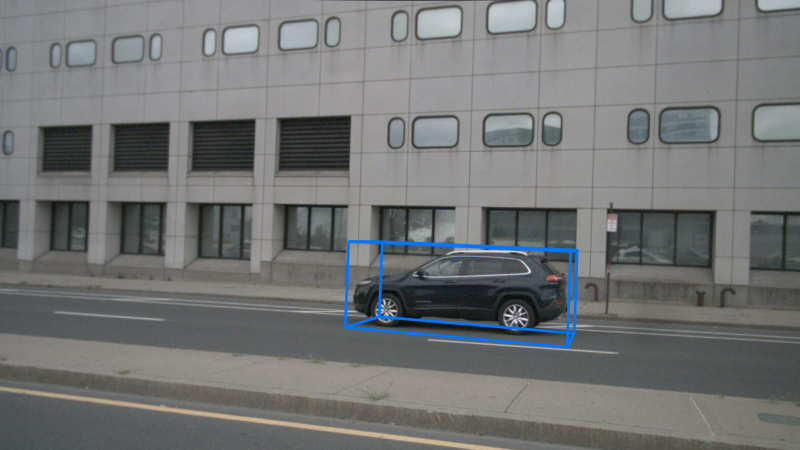} &
    \includegraphics[width=0.32\linewidth]{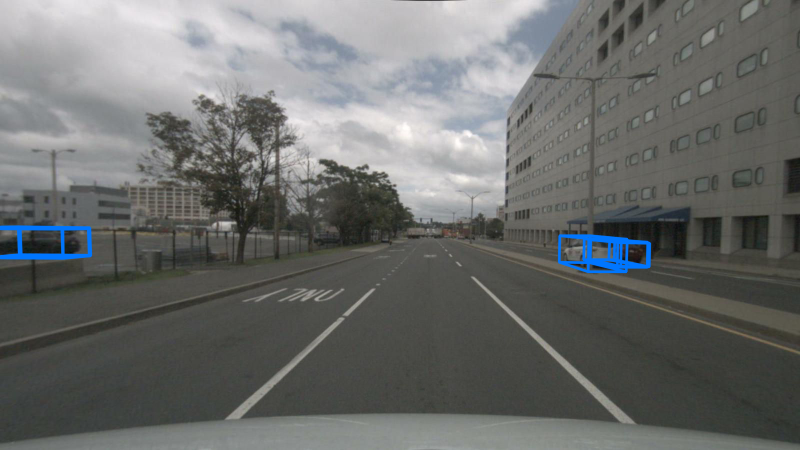} &
    \includegraphics[width=0.32\linewidth]{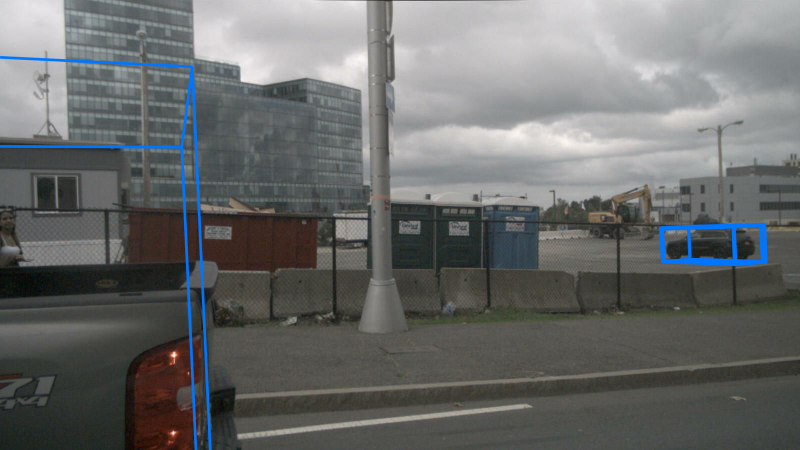} \\
\end{tabular}
a) Scene \textit{n008-2018-08-01-15-16-36-0400\_\_15331512526}.
\\ \vspace{3pt}
\begin{tabular}{ccc}
    \includegraphics[width=0.32\linewidth]{images/nuscenes/n008-2018-09-18-15-12-01-0400__CAM_FRONT_LEFT__1537298104654799.png} &
    \includegraphics[width=0.32\linewidth]{images/nuscenes/n008-2018-09-18-15-12-01-0400__CAM_FRONT__1537298104662404.png} &
    \includegraphics[width=0.32\linewidth]{images/nuscenes/n008-2018-09-18-15-12-01-0400__CAM_FRONT_RIGHT__1537298104670482.png} \\
\end{tabular}
\begin{tabular}{ccc}
    \includegraphics[width=0.32\linewidth]{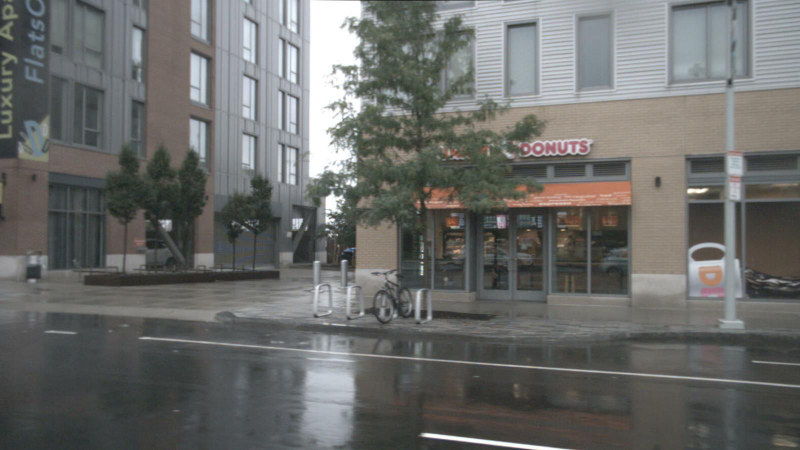} &
    \includegraphics[width=0.32\linewidth]{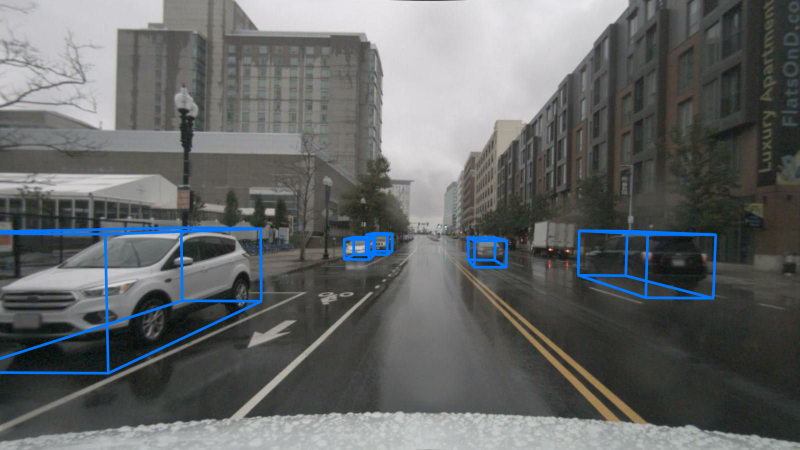} &
    \includegraphics[width=0.32\linewidth]{images/nuscenes/n008-2018-09-18-15-12-01-0400__CAM_BACK_RIGHT__1537298104678113.png} \\
\end{tabular}
b) Scene \textit{n008-2018-09-18-15-12-01-0400\_\_15372981046}.
\caption{Cars detected in the images of two scenes from the validation subset of the nuScenes dataset. The predictions were obtained in multi-view settings. The first two rows correspond to the first scene, and the last two rows correspond to another one. For each scene, the upper row consists of images taken with a front-left, front, and front-right camera (from left to right). The second row contains images taken with a back-left, back, and back-right camera, respectively.}
\label{fig:vis_nuscenes}
\end{figure*}

\begin{figure*}[!ht]
\centering
\setlength{\tabcolsep}{2pt}
\renewcommand{\arraystretch}{0.75}
\begin{tabular}{ccc}
    \includegraphics[width=0.32\linewidth]{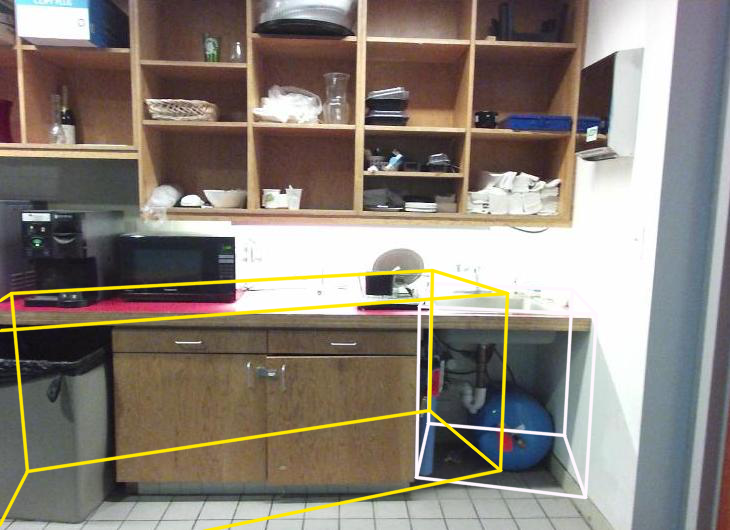} &
    \includegraphics[width=0.32\linewidth]{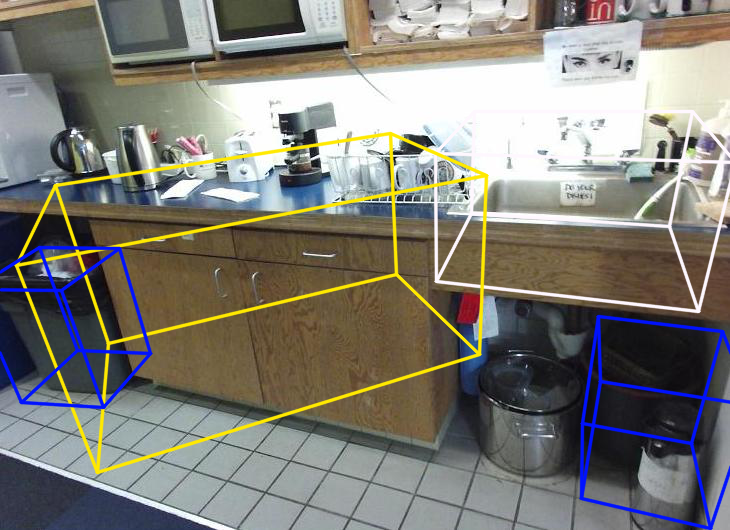} &
    \includegraphics[width=0.32\linewidth]{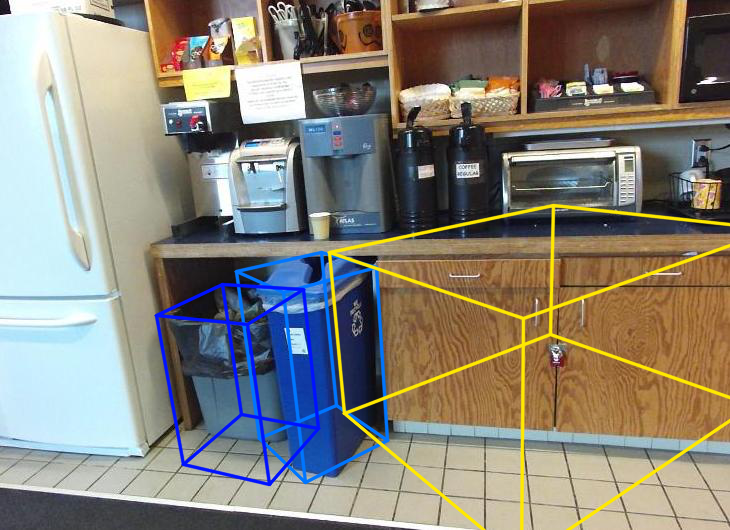} \\
\end{tabular}
\caption{Examples of the detection failures for images from the validation subset of the SUN RGB-D dataset. These examples depict typical error cases: small objects of \textit{sink}, \textit{garbage bin} and \textit{recycle bin} categories are detected quite precisely, but rotation angles for large object such as \textit{cabinet} are estimated poorly.}
\label{fig:failure_sunrgbd}
\end{figure*}

\end{document}